\documentclass[sigconf]{acmart}
\usepackage{amsmath}
\usepackage{bm}
\usepackage{mathrsfs}
\usepackage{amsthm}
\usepackage{amsfonts}
\usepackage{subfigure}
\usepackage{multirow}
\usepackage{url}
\usepackage{array}
  
\usepackage{amssymb}

\usepackage[ruled,vlined]{algorithm2e}

\usepackage{algorithmic}

\AtBeginDocument{%
  \providecommand\BibTeX{{%
    \normalfont B\kern-0.5em{\scshape i\kern-0.25em b}\kern-0.8em\TeX}}}

\copyrightyear{2022} 
\acmYear{2022} 
\setcopyright{acmcopyright}\acmConference[CIKM '22]{Proceedings of the 31st ACM International Conference on Information and Knowledge Management}{October 17--21, 2022}{Atlanta, GA, USA}
\acmBooktitle{Proceedings of the 31st ACM International Conference on Information and Knowledge Management (CIKM '22), October 17--21, 2022, Atlanta, GA, USA}
\acmPrice{15.00}
\acmDOI{10.1145/3511808.3557222}
\acmISBN{978-1-4503-9236-5/22/10}

\begin{document}

\title{A Self-supervised Riemannian GNN \\with Time Varying Curvature for Temporal Graph Learning}



\author{Li Sun}
\email{ccesunli@ncepu.edu.cn}
\affiliation{%
  \institution{North China Electric Power University}
  \city{Beijing}
  \country{China}
}

\author{Junda Ye}
\email{jundaye@bupt.edu.cn}
\affiliation{%
  \institution{Beijing University of Posts and Telecommunications}
  \city{Beijing}
  \country{China}
}

\author{Hao Peng}
\email{penghao@act.buaa.edu.cn}
\affiliation{%
  \institution{Beihang University}
  \city{Beijing}
  \country{China}
}

\author{Philip S. Yu}
\email{psyu@uic.edu}
\affiliation{%
  \institution{University of Illinois at Chicago}
  \city{Chicago}
  \state{IL}
  \country{USA}
}

% \author{Li Sun$^1$, Junda Ye$^2$, Hao Peng$^3$, Philip S. Yu$^4$}
% \affiliation{%
%  $^1$School of Control and Computer Engineering, North China Electric Power University\\
% $^2$School of Computer Science, Beijing University of Posts and Telecommunications\\
% $^3$Beijing Advanced Innovation Center for Big Data and Brain Computing, Beihang University\\
%   $^4$Department of Computer Science, University of Illinois at Chicago
%   }
% \email{ccesunli@ncepu.edu.cn, jundaye@bupt.edu.cn, penghao@act.buaa.edu.cn, psyu@uic.edu}


%\renewcommand{\shortauthors}{Anonymous Author(s)}

\begin{abstract}
Representation learning on temporal graphs has drawn considerable research attention owing to its fundamental importance in a wide spectrum of real-world applications. 
Though a number of studies succeed in obtaining time-dependent representations, 
%the limitations of prior works are still significant.
it still faces significant challenges.
On the one hand, most of the existing methods restrict the embedding space with a certain curvature. However, the underlying geometry in fact shifts among the positive curvature hyperspherical, zero curvature Euclidean and negative curvature hyperbolic spaces in the evolvement over time.
On the other hand, these methods usually require abundant labels to learn temporal representations, and thereby notably limit their wide use in the unlabeled graphs of the real applications. 
To bridge this gap, we make the first attempt to study the problem of \emph{self-supervised temporal graph representation learning in the general Riemannian space}, supporting the time-varying curvature to shift among hyperspherical, Euclidean and hyperbolic spaces.
%in Riemannian space, so that we can learn temporal representations encoding graph evolvement without external guidance. 
In this paper, we present a novel self-supervised Riemannian graph neural network (\textsc{Self}$\mathcal{R}$\textsc{GNN}).
% with the unified formalism in the Riemannian space.
%The evolvement of temporal graph is naturally interpreted as the curvature of the embedding space evolves over time in the language of Riemannian geometry.
Specifically, we design a curvature-varying Riemannian GNN with a theoretically grounded time encoding, 
and formulate a functional curvature over time to model the evolvement shifting among the positive, zero and negative curvature spaces.
%so that we can leverage the notion of curvature to model the evolvement of temporal graphs. 
To enable the self-supervised learning, we propose a novel reweighting self-contrastive approach, exploring the Riemannian space itself without augmentation, and propose
an edge-based self-supervised curvature learning with the Ricci curvature.
%leverage the Ricci curvature to boost the curvature learning of Riemannian space.
% to address the first limitation, we propose a curvature-aware Riemannian graph neural network with unified formalism to model the evolvement of temporal graphs over time.
% % To address the second limitation, we present a Riemannian self-supervised learning approach to obtain the temporal representations without external guidance.
% To enable the self-supervised learning, we propose the Riemannian self-contrastive learning to explore the rich information in the Riemannian space of temporal graphs without the effort for data augmentation, and leverage the Ricci curvature to boost the curvature learning of Riemannian space.
Extensive experiments show the superiority of \textsc{Self}$\mathcal{R}$\textsc{GNN}, and moreover, the case study shows the time-varying curvature of temporal graph in reality.
%that the curvature of temporal graphs trends to evolve over time.
\end{abstract}




\vspace{-0.2in}



\maketitle

% \begin{figure}[h]
%  \vspace{-0.45in}
% \centering 
% \subfigure[Network in May, 1998]{
% \includegraphics[width=0.49\linewidth]{c2}}
% \hspace{-0.02\linewidth}
% \subfigure[Network in May, 2000]{
% \includegraphics[width=0.49\linewidth]{c3}}
%  \vspace{-0.2in}
% \caption{Motivated Example.  The underlying space of the citation network \cite{2019Variational} evolves from positive curvature hyperspherical (1998) to negative curvature hyperbolic space (2000). Corresponding curvatures are learned from data. (Refer to Case Study for details.)
% %(Curves linking the nodes are \emph{geodesics}, manifesting the geometry.)
% }
%  \vspace{-0.3in}
% \label{example}
% \end{figure}
% {}

\section{Introduction}

%Graphs are ubiquitous, describing the relationship among objects in real-world scenarios.
%, such as friendship relation among people and citation relation among papers.
Graph representation learning is now becoming the de facto standard when dealing with the ubiquitous graph data \cite{hamilton2017inductive,DeepWalk,tang2015line}.
%\cite{hamilton2017inductive,DeepWalk,tang2015line}.
%and has achieved great success over a wide spectrum of graph mining tasks, e.g., node classification \cite{hamilton2017inductive} and link prediction \cite{DeepWalk,tang2015line}.
In the literature, the vast majority of graph representation learning methods \cite{kipf2016semi,velickovic2018graph,zhu2018dvne} consider the static setting
%, which embed a static graph into a low-dimensional representation space \footnote{We use representation space and embedding space interchangeably in this paper.} 
with the structure of graphs frozen in time.
In reality, an abundance of graphs are temporal in nature and constantly evolving over time, referred to as \emph{temporal graphs}.
For instance, new interactions (edges) constantly  arrive and the structure of graph changes in social networks, citation networks, e-commerce networks and the World Wide Web \cite{zuo2018embedding,zhou2018dynamic}. 
Naive application of the static methods on temporal graphs fails to capture the temporal information, and ignoring the temporal information usually leads to questionable inference \cite{Xu2020inductive,2019Variational}. 
%, e.g., static methods may mistakenly utilize future information for predicting past interactions as the evolving constraints are disregarded.
Thus, the representation learning on temporal graphs has drawn increasing attention in recent years \cite{abs-2006-10637,LiuHYD21,BianKDD19,DuWSLW18}. {}

% VGRNN \cite{2019Variational,pareja2019evolvegcn}.
% For the continuous-time methods, temporal random walks \cite{CTDNE,liu2020fine} have shown to be effective, and the recent CAWNet \cite{CAWNet,DyRep,Xu2020inductive} is based on the causual anonymous walks.
% %to learn the temporal representations inductively
% Temporal point process \cite{TREND,zuo2018embedding,HeteroHawkes} is another important tool, e.g., DyRep  considers an additional hop of interactions for further expressiveness. 
% Recently, GNN-based models have also emerged to deal with continuous time, e.g., TGAT \cite{} extends GAT \cite{velickovic2018graph} to the temporal graphs. 
To date, a series of temporal graph representation learning methods have been designed to output time-dependent representations \cite{AggarwalS14,KazemiGJKSFP20}, which can be roughly divided into two main categories: discrete-time methods and continuous-time methods. 
The discrete-time methods frame the temporal graphs into a sequence of snapshots, and recurrent architectures are frequently employed \cite{2019Variational,pareja2019evolvegcn}. A major drawback of discrete-time methods is that the appropriate granularity for temporal discretization is often subtle.
%In this paper, we summarize the limitations of the prior graph representation learning works in three folds
In contrast, the continuous-time methods \cite{CAWNet,DyRep,Xu2020inductive}, which directly integrate temporal information into the representation learning, are able to model the temporal graphs with a finer granularity.
Despite the success of prior works,  representation learning on temporal graphs still faces significant challenges.
% we have also observed that the existing methods for temporal graphs still suffer from significant limitations in two major perspectives, which are described as follows:

%(positive/0/negative)
\emph{\textbf{ Challenge 1}: Embedding Space Supporting Time-varying Curvature.}
To the best of our knowledge, existing studies restrict the embedding space in a Riemannian space of certain curvature.
%either the traditional zero-curvature Euclidean space or the negative-curvature hyperbolic space.
In the literature, the vast majority of temporal graph models \cite{2019Variational,pareja2019evolvegcn, CAWNet,DyRep,Xu2020inductive} work with traditional zero-curvature Euclidean space, 
% e.g., Euclidean distance and inner product, and thereby explicitly or implicitly embeds a temporal graph into Euclidean space. 
% In  recent years,
% Recent advances of network science show that  hyperbolic space is well-suited to model the graphs with latent hierarchical or tree-like structures \cite{chen2013hyperbolicity,krioukov2010hyperbolic}, 
and it is not until very recently a few graph neural networks \cite{HTGN,HVGNN} for temporal graphs  are proposed in the negative-curvature hyperbolic space.
% However, the underlying space of the temporal graphs is usually not restricted in a certain curvature, and in reality, the curvature evolves as the structure of temporal graph evolves over time \cite{ravasz2003hierarchical,papadopoulos2012popularity}, which is also evidenced in our case study.
In fact, rather than restricted in a certain curvature, the curvature of the underlying space varies as the temporal graph evolves over time \cite{ravasz2003hierarchical,krioukov2010hyperbolic}. 
Even more challenging, the time-varying curvature actually shifts among the positive-curvature hyperspherical, zero-curvature Euclidean and negative-curvature hyperbolic spaces in the evolvement over time \cite{papadopoulos2012popularity}.
% ., as shown in the example of Fig. \ref{example}, and we will provide further discussion in the case study.
%and thus the curvature of the embedding space matching its geometry is usually covered.
%Therefore, it calls for a promising principled approach with a unified formalism in the Riemannian space of arbitrary curvature to model the evolvement of temporal graphs.
Therefore, it calls for a promising approach in the general Riemannian space with time-varying curvature to model the evolvement shifting among the Riemannian spaces of various curvatures.
%, i.e., positive, zero, and negative. 
%(positive/0/negative).

\emph{\textbf{Challenge 2}: Self-supervised Learning  for Temporal graph in Riemannian space.}
Most of the learning methods \cite{HTGN,HVGNN,CAWNet,pareja2019evolvegcn,JOIDE} require abundant labels to learn the time-dependent representations. 
Labels are usually scarce in real applications, and undoubtedly, labeling graphs is expensive, either manual annotation or paying for permission, and is even impossible to acquire because of the privacy policy. 
Hence, representation learning on temporal graphs without labels is more preferable,
and fortunately, self-supervised learning \cite{DBLP:journals/corr/abs-2006-08218,ChenK0H20} has recently emerged as a principled way by exploring the similarity of data themselves without external annotations.
In the literature, though the self-supervised learning for static graphs is being extensively studied \cite{VelickovicFHLBH19,HassaniA20,QiuCDZYDWT20}, the effort for temporal graph is still limited.
Recently,  \citet{DDGCL} propose a self-supervised learning method for temporal graph in the traditional Euclidean space.
However, it cannot be applied to the general Riemannian space owing to essential distinction in geometry.
That is, to the best of our knowledge, self-supervised learning for temporal graphs still remains open, especially for the general Riemannian space with time-varying curvature.

To address the aforementioned challenges, we propose to  study the problem of \emph{self-supervised temporal graph representation learning in the general Riemannian space} for the first time so as to learn temporal representations modeling the graph evolvement over Riemannian spaces of various curvatures without external guidance.

In this paper, we propose a novel  \underline{Self}-supervised \underline{R}iemannian \underline{G}raph \underline{N}eural \underline{N}etwork, referred to \textsc{Self}$\mathcal{R}$\textsc{GNN}, for the representation learning on temporal graphs.
The evolvement of temporal graph  is naturally described as the time-varying curvature of the embedding space in the language of Riemannian geometry.
Consequently, we first propose a curvature-varying Riemannian graph neural network, in which we formulate a time encoding of arbitrary curvature to capture the temporal information, and further prove that the encoding function is translation invariant in time.
Then, we formulate a functional curvature over time to model the temporal evolvement over Riemannian spaces of various curvatures from hyperspherical to Euclidean and hyperbolic spaces.
%Then, we leverage the notion of curvature to model the evolvement of temporal graphs in Riemannian space. 
%propose a curvature-aware Riemannian graph neural network with unified formalism to model the evolvement of temporal graphs over time.
To enable its self-supervised learning, we propose a Riemannian reweighted self-contrastive approach to learn temporal representations in the absence of labels. 
Specifically, we introduce a novel self-augmentation underpinned by the functional curvature to get rid of introducing new graphs, and formulate a reweighted contrastive objective that reweights the negative samples without sampling bias.
In addition, we propose an edge-based self-supervised curvature learning with the well-defined Ricci curvature, completing the self-supervised learning of \textsc{Self}$\mathcal{R}$\textsc{GNN}. 

\noindent \textbf{Contributions.} Noteworthy contributions are summarized below:
\begin{itemize}
  \vspace{-0.03in}
  \item \emph{Problem.} To the best of our knowledge, we make the first attempt on formulating the representation learning problem for temporal graphs in the general Riemannian space, supporting time-varying curvature to shift among hyperspherical, Euclidean and hyperbolic spaces in the evolvement over time.
  \item \emph{Methodology.} We propose the novel \textsc{Self}$\mathcal{R}$\textsc{GNN}, in which the curvature-varying Riemannian GNN and its functional curvature over time are designed to model the evolvement in the general Riemannian space of various curvatures, and the Riemannian reweighting self-contrastive approach is proposed to enable its self-supervised learning.
  \item \emph{Experiments.} Extensive experiments on real-world datasets show that \textsc{Self}$\mathcal{R}$\textsc{GNN} even outperforms the state-of-the-arts supervised methods, and the case study shows the time-varying curvature of temporal graph in reality.
  %The ablation study gives further insights into how each proposed component contributes to the success of the model, and 
 %and the case study shows that the curvature of temporal graphs trends to evolve over time.
  %shows that temporal graphs evolve over the Riemannian space of different curvatures.

\end{itemize}

\vspace{-0.1in}
\section{Preliminaries and Problem}

In this section, we first introduce the fundamentals of Riemannian geometry and the notation of curvature, and then formulate the problem of temporal graph learning in general Riemannian Space.

\vspace{-0.07in}
\subsection{Preliminaries}

\noindent{\textbf{Riemannian Geometry.}} It provides an elegant mathematical framework to study the geometry beyond Euclid.
The fundamental object in Riemannian geometry is a smooth \emph{manifold} $\mathcal M$, which generalizes the notion of the surface to higher dimensions.
Each point $\boldsymbol x \in \mathcal M$ associates with a \emph{tangent space} $\mathcal T_{\boldsymbol x}\mathcal M$, the first order approximation of $\mathcal M$ around $\boldsymbol x$.
On the tangent space of $\boldsymbol x$, the \emph{Riemannian metric}, $g_{\boldsymbol x} (\cdot, \cdot) : \mathcal T_{\boldsymbol x}\mathcal M  \times \mathcal T_{\boldsymbol x}\mathcal M \to \mathbb R$, defines an inner product so that geometric notions can be induced.
A \emph{Riemannian manifold} is then defined on the smooth manifold paired with a Riemannian metric, denoted as the tuple $(\mathcal M, g)$. 
The length of the shortest walk connecting two points $\boldsymbol x, \boldsymbol y$ on the Riemannian manifold is called (geodesic) distance $d_\mathcal M(\boldsymbol x, \boldsymbol y)$.
Refer to mathematical materials \cite{Spivak1979,1997Riemannian} for in-depth expositions.

\noindent{\textbf{Curvature on the Manifold.}} In Riemannian geometry, the \emph{constant curvature} $\kappa$ is the notion to measure how a smooth manifold deviates from being flat.
% The Riemannian metric also defines a curvature at each point $\kappa(\mathbf x)$, 
% which determines how the space is curved.
% If the curvature is uniformly distributed,  
% $(\mathcal M, g)$ is called a \emph{constant curvature space} of curvature $\kappa$. 
There are three canonical types of constant curvature space that we can define with respect to its sign: 
the positively curved hyperspherical space $\mathbb S$ ($\kappa>0$), 
the negatively curved hyperbolic space $\mathbb H$ ($\kappa<0$), 
and the flat Euclidean space $\mathbb E$  ($\kappa=0$), which is regarded as a special case.
% of the  Riemannian geometry. 

\noindent{\textbf{Ricci Curvature.}} More specifically, for a point $\boldsymbol x$ on the manifold, and for each pair of linearly independent vectors $\boldsymbol v$ and  $\boldsymbol u$ in $\mathcal T_{\boldsymbol x}\mathcal M$, the sectional curvature at $\boldsymbol x$ is defined on the surface spanned by the exponential map of $\boldsymbol v$ and $\boldsymbol u$, encoding the local geometry around ${\boldsymbol x}$.
Given a tangent vector $\boldsymbol v$ at $\boldsymbol x$, if we average the sectional curvatures at $\boldsymbol x$ over a set of orthonormal vectors, we obtain the \emph{Ricci curvature}, from which the \emph{constant curvature} $\kappa$ of the Riemannian space can be induced \cite{1997Riemannian}.
In this paper, we leverage the coarse Ricci curvature \cite{RcciOlli} to provide supervision signal for the proposed model.

\begin{figure*}
\centering
       \vspace{-0.1in}
    \includegraphics[width=0.95\linewidth]{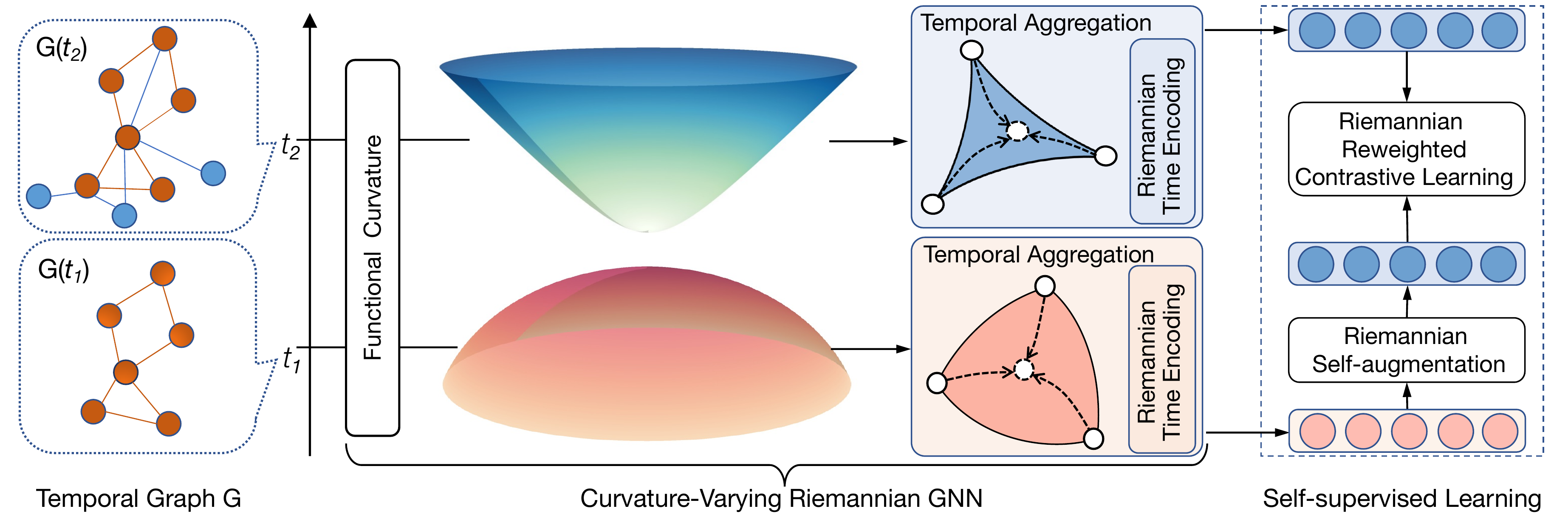}
    \vspace{-0.1in}
    \caption{Overall architecture of  \textsc{Self}$\mathcal{R}$\textsc{GNN}. We first design a curvature-varying $\mathcal{R}$GNN shifting among the Riemannian spaces of various curvatures, e.g., positive hyperspherical (\textcolor{orange}{orange}) and negative hyperbolic (\textcolor{blue}{blue}) spaces, and then enable its self-supervised learning with the self-augmentation and reweighted contrastive objective without introducing new graphs.}
    \label{illu}
        \vspace{-0.1in}
\end{figure*}

\subsection{Problem Formulation}

In this paper, we consider the temporal graph that evolves over time, and utilize the interactions with temporal information (timestamped edges) to model the  temporal graphs  with fine granularity. 

% \newtheorem*{def1}{Definition 1 (Temporal Graph)} 
% \begin{def1}
% A temporal graph, denoted as the tuple $\mathcal G = (\mathcal V, \mathcal E, \boldsymbol X, T)$, is defined on a set of nodes $\mathcal V = \{v_1,  v_2,  \cdots, v_N\}$, a set of edges $\mathcal E = \{(v_i,  v_j)\}$, a time domain $T$ and a feature matrix $\boldsymbol X \in \mathbb R^{N \times F}$. The number of nodes is $N=|V|$, and each node $v_i$ is associated with a feature vector recorded in the corresponding row in $\boldsymbol X$, denoted as $\boldsymbol x_i \in \mathbb R^F$. Each edge $(v_i,  v_j)$ is attached with a timestamp $t_{ij} \in T$, describing the interaction between $v_i \in \mathcal V$ and $v_j \in \mathcal V$ is at time $t_{ij} \in T$.
% \end{def1}

\newtheorem*{def1}{Definition 1 (Temporal Graph)} 
\begin{def1}
A temporal graph, denoted as the tuple $\mathcal G = (\mathcal V, \mathcal E, \boldsymbol X, T)$, is defined on a set of nodes $\mathcal V = \{v_1,  v_2,  \cdots, v_N\}$, a set of timestamped edges $\mathcal E = \{(v_i,  v_j, t_l)\}$, a time domain $T$ and a feature matrix $\boldsymbol X \in \mathbb R^{N \times F}$. The number of nodes is $|\mathcal V|$, and each node $v_i$ is associated with a feature vector recorded in the corresponding row in $\boldsymbol X$, denoted as $\boldsymbol x_i \in \mathbb R^F$. A timestamped edge $(v_i,  v_j, t_l)$ describes the temporal interaction between nodes $v_i \in \mathcal V$ and $v_j \in \mathcal V$ at time $t_l \in T$.
\end{def1}
\vspace{-0.03in}
In temporal graphs, for a given node, the members of its neighborhood is time-dependent with the new arrival of edges, which is different from that of the static setting in essence. Hence, we define the neighborhood in temporal graphs as temporal neighborhood, and  give the formal definition as follows:
\vspace{-0.03in}
\newtheorem*{def2}{Definition 2 (Temporal Neighborhood)} 
\begin{def2}
Given a time point $t$, a temporal neighborhood of $v_i$, denoted as $\mathcal N_{t}(v_i)$, is defined as a collection of the nodes linking to $v_i$ with the elder timestamps, i.e., $\mathcal N_{t}(v_i)=\{v_j | \ (v_i, v_j, t_l) \in \mathcal E \land t_{ij} \le t\}$.
\end{def2}

% \newtheorem*{def2}{Definition 2 (Temporal Neighborhood)} 
% \begin{def2}
% Given a time point $t$, the temporal neighborhood of $v_i$, denoted as $\mathcal N_{t}(v_i)$, is defined as a collection of edges linked to $v_i$ with the elder timestamps, i.e., $\mathcal N_{t}(v_i)=\{(v_i, v_j, t_l) | \ (v_i, v_j, t_l) \in \mathcal E \land t_{l} \le t\}$.
% \end{def2}

With the preliminaries on Riemannian geometry and definitions above, we formulate the studied problem as follows:
\vspace{-0.03in}
\newtheorem*{def3}{Problem Definition (Self-supervised Temporal Graph Representation Learning in the General Riemannian Space)} 
\begin{def3}
Given a temporal graph $\mathcal G = (\mathcal V, \mathcal E, \boldsymbol{X}, T)$,  the problem  is to find an encoding function $\Phi: \mathcal V \to \mathcal M^{d, \kappa}$  so that, for each node $v_i$, we can infer the representation $\boldsymbol h_i(t)$ at time $t$ in the general Riemannian space with time-varying curvature without any external guidance, encoding the evolvement of the temporal graph over time.
\end{def3}

In other words, we are interested in designing a novel encoding function for temporal graphs that i) can model the graph evolvement shifting among the positive-curvature hyperspherical, zero-curvature Euclidean and negative-curvature hyperbolic spaces over time, and ii) is endowed with the self-supervised learning ability.

%learn temporal representations in the general Riemannian space to

\vspace{-0.05in}
\section{Methodology}
In this section, we propose a novel \underline{Self}-supervised \underline{R}iemannian \underline{G}raph \underline{N}eural \underline{N}etwork (\textsc{Self}$\mathcal{R}$\textsc{GNN}) for the temporal graph learning in the general Riemannian space with time-varying curvature. 
We illustrate the overall architecture of \textsc{Self}$\mathcal{R}$\textsc{GNN} in Figure 2. 
As sketched in Figure 2,  we first design a curvature-varying Riemannian graph neural network, modeling the evolvement shifting among the Riemannian spaces of various curvatures over time, 
and then propose a novel Riemannian self-supervised learning approach, obtaining temporal representations in the absence of external guidance. 
%Next, we are to elaborate on the components of \textsc{Self}$\mathcal{R}$\textsc{GNN} in the following subsections.
Next, we will elaborate on each component in the following subsections, respectively.

First of all, we introduce the Riemannian manifolds we use in this paper before we construct \textsc{Self}$\mathcal{R}$\textsc{GNN} on them.

\noindent \textbf{Riemannian manifolds}: We opt for the hyperboloid (Lorentz) model for hyperbolic space and the corresponding hypersphere model for hyperspherical space with the unified formalism, owing to the numerical stability and closed form expressions \cite{HGCN,ZhangWSLS21}.
Specifically, we have a  $d$-dimensional manifold of curvature $\kappa$,
\vspace{-0.05in}
\begin{equation}
\mathcal M^{d, \kappa}=\left\{
\begin{array}{lll}
\mathbb S^{d, \kappa}&=\{\boldsymbol{x} \in \mathbb R^{d+1} :  \langle \boldsymbol{x}, \boldsymbol{x} \rangle_2= \frac{1}{\kappa}\} & \text{for} \  \kappa>0, \\  
 \mathbb E^{d}&=\{\boldsymbol{x} \in \mathbb R^d\} & \text{for} \  \kappa=0, \\  
\mathbb H^{d, \kappa}&=\{\boldsymbol{x} \in \mathbb R^{d+1}:  \langle \boldsymbol{x}, \boldsymbol{x} \rangle_\mathcal L= \frac{1}{\kappa}\}& \text{for} \  \kappa<0,   
\end{array}\right.
\label{manifold}
\end{equation}
\noindent where $\langle \cdot, \cdot \rangle_2$ and $\langle \cdot, \cdot \rangle_\mathcal L$ denote the standard inner product and Minkowski inner product on $\mathbb R^{d+1}$, respectively, and the Minkowski inner product is defined as 
\vspace{-0.05in}
\begin{equation}
\langle \boldsymbol{x}, \boldsymbol{y} \rangle_\mathcal L=\boldsymbol{x}^{\top} \operatorname{diag}(-1,1, \cdots, 1) \boldsymbol{y}.
\end{equation}
The \emph{origin} of the manifold $(\sqrt{\frac{1}{|\kappa|}}, 0, \cdots, 0)$ is denoted as $\mathcal O \in \mathcal M^{d,\kappa}$.
% i.e., $(\sqrt{\frac{1}{-\kappa}}, 0, \cdots, 0)$ for the hyperboloid and $(\sqrt{\frac{1}{\kappa}}, 0, \cdots, 0)$ for the hypersphere.

\vspace{-0.07in}
\subsection{Curvature-varying Riemannian GNN}

%In the Riemannian geometry, the evolvement of temporal graphs over time is naturally described as the curvature evolvement of embedding space, and thereby
We propose a novel curvature-aware Riemannian GNN ($\mathcal{R}$\textsc{GNN}) with a theoretically grounded time encoding to define the encoding function $\Phi$ above, in which we formulate a functional curvature over time  to model temporal evolvement shifting among Riemannian spaces of various curvatures (positive, zero and negative).
This novel idea distinguishes us with  the vast majority of existing studies that restrict the embedding space in a Riemannian space of certain curvature  with an inductive bias given in prior.

%hat we formulate a unified formalism in the Riemannian of arbitrary curvature that can model the evolvement of a temporal graph as its structure evolves over time, in contrast to

\begin{table}
  %\scriptsize
\centering
\caption{Summary of the operations with unified formalism.}
\vspace{-0.1in}
\begin{tabular}{|l|c|}
\hline
\textbf{Operation}  & \textbf{Unified formalism in $\mathcal M^{d, \kappa}$}\\
\hline
Distance Metric
&
$
d_{\mathcal M}(\boldsymbol{x}, \boldsymbol{y})=\frac{1}{\sqrt{|\kappa|}} \cos^{-1}_{\kappa}\left(|\kappa|\langle\boldsymbol{x}, \boldsymbol{y}\rangle_{\kappa}\right)
$\\
\hline
% Exponential Map & 
% $exp_{\boldsymbol{x}}^{\kappa}(\boldsymbol{v})=\boldsymbol{x}+\boldsymbol{v}$ 
% &
% $
% exp _{\boldsymbol{x}}^{\kappa}(\boldsymbol{v})=\cos_{\kappa}\left(\sqrt{|\kappa|}\|\boldsymbol{v}\|_{\kappa}\right) \boldsymbol{x}+\sin_{\kappa}\left(\sqrt{|\kappa|}\|\boldsymbol{v}\|_{\kappa}\right) \frac{\boldsymbol{v}}{\sqrt{|\kappa|}\|\boldsymbol{v}\|_{\kappa}}
% $
% \\
% Logarithmic Map & 
% $log_{\boldsymbol{x}}^{\kappa}(\boldsymbol{y})= \boldsymbol{x}-\boldsymbol{y}$
% &
% $
% log _{\boldsymbol{x}}^{\kappa}(\boldsymbol{y})=\frac{\arccos_{\kappa} \left(\kappa\langle\boldsymbol{x}, \boldsymbol{y}\rangle_{\kappa}\right)}{\sin _{\kappa}\left(\arccos_{\kappa}\left(\kappa\langle\boldsymbol{x}, \boldsymbol{y}\rangle_{\kappa}\right)\right)}\left(\boldsymbol{y}-\kappa\langle\boldsymbol{x}, \boldsymbol{y}\rangle_{\kappa} \boldsymbol{x}\right)
% $
% \\
% % Parallel Transport & 
% % $
% % \operatorname{PT}_{\boldsymbol{x} \rightarrow \boldsymbol{y}}^{\kappa}(\boldsymbol{v})=\boldsymbol{v}-\frac{\kappa\langle\boldsymbol{y}, \boldsymbol{v}\rangle_{\kappa}}{1+\kappa \langle\boldsymbol{x}, \boldsymbol{y}\rangle_{\kappa}}(\boldsymbol{x}+\boldsymbol{y})
% % $
% % \\
% \hline
% % Addition & 
% % $
% % \boldsymbol{x} \oplus_{\kappa} \boldsymbol{y}=\frac{\left(1+2 \kappa \boldsymbol{x}^{T} \boldsymbol{y}+K\|\boldsymbol{y}\|^{2}\right) \boldsymbol{x}+\left(1-\kappa || \boldsymbol{x}||^{2}\right) \boldsymbol{y}}{1+2 \kappa \boldsymbol{x}^{T} \boldsymbol{y}+\kappa^{2}|| \boldsymbol{x}||^{2}|| \boldsymbol{v}||^{2}}
% % $\\
Scalar Multiplication 
&
$
r \otimes_{\kappa} \boldsymbol{x}=exp _{\mathcal O}^{\kappa}\left(r \ log_{\mathcal O}^{\kappa}(\boldsymbol{x})\right) 
$\\
Matrix Multiplication 
& 
$
\boldsymbol M \otimes_{\kappa} \boldsymbol{x}=exp _{\mathcal O}^{\kappa}\left(\boldsymbol M \ log_{\mathcal O}^{\kappa}(\boldsymbol{x})\right) 
$\\
Applying Functions 
&
$
f^{\otimes_{\kappa}}(\boldsymbol x)=exp _{\mathcal O}^{\kappa}\left(f\left(log_{\mathcal O}^{\kappa}(\boldsymbol x)\right)\right)
$\\
\hline
\end{tabular}
\label{tab:ops}
\vspace{-0.1in}
\end{table}

\vspace{-0.03in}
\subsubsection{Time Encoding of Arbitrary Curvature}

%The temporal information plays a crucial role in the representation learning on temporal graphs.
Interaction time between nodes (timestamps) records the graph evolvement specifically.
To tackle with timestamps, we propose a novel time encoding function of arbitrary curvature $\varphi^\kappa: t \to \boldsymbol t^\kappa \in \mathcal M^{d,\kappa}$ which maps a time point $t \in T$ to a vector $\boldsymbol t^\kappa$ in Riemannian space of curvature $\kappa$, so that the temporal information is transformed as a Riemannian feature compatible with the graph convolution.
We begin the elaboration with the simple and special case, Euclidean time encoding.

\noindent\textbf{Euclidean Time Encoding.} 
In a nutshell, the Euclidean time encoding defines a generic function $\varphi^0: t \to \boldsymbol t^0$, where $ t \in T$ and $\boldsymbol t^0 \in \mathbb R^d$.
Stemming from the design of random Fourier features, a Bochner-type (trigonometric) encoding function can be derived  as
\begin{equation}
\varphi^0(t)=\sqrt{\frac{1}{d}}\left[\cos \left(\omega_{1} t\right), \sin \left(\omega_{1} t\right), \ldots, \cos \left(\omega_{d} t\right), \sin \left(\omega_{d} t\right)\right],
\label{EucEncoding}
\end{equation}
\noindent parameterized by the $\omega$'s \cite{Xu2020inductive}. According to the Bochner’s Theorem, the encoding function is translation-invariant in time. Specifically, there exists a function $\psi^{E}(\cdot)$ on $\mathbb R$ so that the induced kernel $\mathcal K^{E}(t_i, t_j) = \langle  \varphi^0(t_i) , \varphi^0(t_j) \rangle _{2}$ can be expressed as $\mathcal K^{E}(t_i, t_j) = \psi^{E}(t_i- t_j)$, i.e., we have the following equation holds for any $t_0$,
\begin{equation}
\mathcal K^{E}(t_i, t_j) =\mathcal K^{E}(t_0-t_i,t_0- t_j) = \psi^{E}(t_i- t_j).
\end{equation}

\noindent\textbf{Riemannian Time Encoding.}
Based on the Euclidean encoding above, we formulate a Riemannian time encoding of arbitrary curvature with the aid of exponential map, and further \emph{prove that the translation invariant property also holds for the proposed formulation}.

\newtheorem*{prop1}{Proposition 1 (Exponential and Logarithmic Maps)} 
\begin{prop1}
For the points on the manifold of curvature $\kappa$,   $\boldsymbol x, \boldsymbol y \in \mathcal M^{d, \kappa}$, and $\boldsymbol{v}$ in the tangent space of $\boldsymbol x$, $\boldsymbol v \in \mathcal T_{\boldsymbol x} \mathcal M^{d, \kappa}$, such that $\boldsymbol x \neq \boldsymbol y$ and $\boldsymbol v \neq \boldsymbol 0$,
the exponential map $exp_{\boldsymbol x}(\boldsymbol v): \ \mathcal T_{\boldsymbol{x}} \mathcal M^{d, \kappa} \to \mathcal M^{d, \kappa}$ at $\boldsymbol x$ projects the vector $\boldsymbol v$ of its tangent space onto the manifold $\mathcal M^{d, \kappa}$,
$$
exp _{\boldsymbol{x}}^{\kappa}(\boldsymbol{v})=\cos_{\kappa}\left(\sqrt{|\kappa|}\|\boldsymbol{v}\|_{\kappa}\right) \boldsymbol{x}+\sin_{\kappa}\left(\sqrt{|\kappa|}\|\boldsymbol{v}\|_{\kappa}\right) \frac{\boldsymbol{v}}{\sqrt{|\kappa|}\|\boldsymbol{v}\|_{\kappa}}.
$$
The logarithmic map $log_{\boldsymbol x}^\kappa(\boldsymbol y): \  \mathcal M^{d, \kappa} \to \mathcal T_{\boldsymbol{x}} \mathcal M^{d, \kappa}$ at $\boldsymbol x$ projects the vector $\mathbf y \in \mathcal M$ back to the tangent space $\mathcal T_\mathbf x\mathcal M^{d, \kappa}$,
$$
log _{\boldsymbol{x}}^{\kappa}(\boldsymbol{y})=\frac{\cos^{-1}_{\kappa} \left(\kappa\langle\boldsymbol{x}, \boldsymbol{y}\rangle_{\kappa}\right)}{\sin _{\kappa}\left(\cos^{-1}_{\kappa}\left(\kappa\langle\boldsymbol{x}, \boldsymbol{y}\rangle_{\kappa}\right)\right)}\left(\boldsymbol{y}-\kappa\langle\boldsymbol{x}, \boldsymbol{y}\rangle_{\kappa} \boldsymbol{x}\right)
$$
\end{prop1}

\noindent \emph{Remarks}: We summarize all the necessary operations for this paper in Table \ref{tab:ops} with the curvature-aware definition of trigonometrics, e.g., $\cos_\kappa(\cdot)=\cosh(\cdot)$ if $\kappa<0$ and $\cos_\kappa(\cdot)=\cos(\cdot)$ if $\kappa>0$.

%e can obtain the closed form expressions of  exponential and logarithmic maps on $\mathcal M^{d, \kappa}$ with unified formalism as shown , in which we

We derive the Riemannian time encoding of arbitrary curvature as follows. Specifically, we first augment $\boldsymbol{t}^0$ as $\bar{\boldsymbol{t}^0}=[0  || \boldsymbol{t}^0]$, where $[ \cdot || \cdot ]$ denotes the concatenation of vectors, and it is easy to check that the augmented encoding resides in the tangent space of the origin $\mathcal O$ of the Riemannian manifold, i.e., $\langle \mathcal O, \bar{\boldsymbol{t}^0}\rangle_\kappa=0$ holds for any $\boldsymbol{t}^0$.
Then, we project the tangent vector $ \bar{\boldsymbol{t}^0}$ via the exponential map at the origin $exp^\kappa_\mathcal O\left(\cdot\right)$ to obtain $\boldsymbol{t}^\kappa$ at Riemannian manifold of curvature $\kappa$, yielding the Riemannian time encoding as follows:
%$ \boldsymbol{t}^\kappa =\pi^\kappa(\bar{\boldsymbol{t}^0}) =$
\vspace{-0.05in}
\begin{equation} 
\boldsymbol{t}^\kappa = \left[\frac{ \cos_\kappa \left(\sqrt{|\kappa|}\|  \bar{\boldsymbol{t}^0}  \|_{\kappa}\right) }{\sqrt{|\kappa|}},
\frac{ \sin_\kappa \left(\sqrt{|\kappa|}\|  \bar{\boldsymbol{t}^0}  \|_{\kappa}\right) }{ \sqrt{|\kappa|}\|  \bar{\boldsymbol{t}^0}  \|_{\kappa} } \boldsymbol{t}^0 \right],
\vspace{-0.05in}
\end{equation}
where $\boldsymbol{t}^0$ is the Euclidean encoding.
$\|  \bar{\boldsymbol{t}^0}  \|_{\kappa}=\sqrt{\langle [0, \boldsymbol{t}^0], [0, \boldsymbol{t}^0]\rangle_\mathcal L}=\|  \boldsymbol{t}^0  \|_2$ when $\kappa <0$,
while $\|  \bar{\boldsymbol{t}^0}  \|_{\kappa}=\sqrt{\langle [0, \boldsymbol{t}^0], [0, \boldsymbol{t}^0]\rangle_2}=\|  \boldsymbol{t}^0  \|_2$ when $\kappa >0$. 
Note that, $\|  \bar{\boldsymbol{t}^0}  \|_{\kappa}=\|  \boldsymbol{t}^0  \|_2=1$.
Therefore, we give the unified formulation of the time encoding function as follows
\vspace{-0.05in}
\begin{equation}
 \varphi^\kappa\left(t\right) = \left[\frac{ \cos_\kappa \left(\sqrt{|\kappa|}\right) }{\sqrt{|\kappa|}},
\frac{ \sin_\kappa \left(\sqrt{|\kappa|}\right) }{ \sqrt{|\kappa|}} \varphi^0\left(t\right)\right],
\vspace{-0.05in}
\end{equation}
for any $t \in T$, where $\varphi^0\left( \cdot \right)$ is defined in Eq. (\ref{EucEncoding}).

Moreover, we prove that the Riemannian time encoding is translation invariant in time as its Euclidean counterpart.

\newtheorem*{prop2}{Proposition 2 (Translation Invariant Property of Riemannian Time Encoding)} 
\begin{prop2}
Given the Riemannian time encoding of arbitrary curvature  $\varphi^\kappa\left( \cdot \right)$, 
the induced Riemannian kernel $\mathcal K^{R}(t_i, t_j) = \langle  \varphi^\kappa(t_i) , \varphi^\kappa(t_j) \rangle _{\kappa}$ is translation invariant over time in domain $T$, i.e, there exists a function $\psi^{R}(\cdot)$ so that 
$$\mathcal K^{R}(t_i, t_j) = \psi^{R}(t_i-t_j).$$
\end{prop2}

\begin{proof}
We provide the idea and key equations only due to the limit of space. We prove the translation invariant property of the induced Riemannian kernel by proving the existence of the function $\psi^{R}(\cdot)$ regardless of the sign of $\kappa$. 
Specifically, we have
\vspace{-0.05in}
\begin{equation}
\begin{aligned}
\mathcal K^{R}(t_i,t_j) &= \langle \boldsymbol{t}^\kappa_i, \boldsymbol{t}^\kappa_j \rangle_\kappa \\
&=-\frac{\left( \sin_\kappa \left(\sqrt{|\kappa|}\right) \right)^2}{ |\kappa|} \langle \boldsymbol{t}^0_i, \boldsymbol{t}^0_j \rangle_\kappa + \frac{\left( \cos_\kappa \left(\sqrt{|\kappa|}\right) \right)^2}{| \kappa|}  \\
&= -\frac{\left( \sin_\kappa \left(\sqrt{|\kappa|}\right) \right)^2}{ |\kappa|}\mathcal K^{E}(t_i,t_j) + \frac{\left( \cos_\kappa \left(\sqrt{|\kappa|}\right) \right)^2}{ |\kappa|} \\
&= \psi^R(t_i - t_j),
\end{aligned}
\vspace{-0.05in}
\end{equation}
and $\psi^R=g \circ \psi^E$.
% \begin{equation}
% A_1=-\frac{\left( \sin_\kappa \left(\sqrt{-\kappa}\right) \right)^2}{ \kappa},  \ \ B_1=\frac{\left( \cos_\kappa \left(\sqrt{-\kappa}\right) \right)^2}{ \kappa}.
% \end{equation}
% Similarly, when $\kappa >0$,
% \begin{equation}
% \begin{aligned}
% \mathcal K^{R}(t_i,t_j) &= \langle \boldsymbol{t}^\kappa_i, \boldsymbol{t}^\kappa_j \rangle_2 \\
% &=\frac{\left( \sin_\kappa \left(\sqrt{\kappa}\right) \right)^2}{ \kappa} \langle \boldsymbol{t}^0_i, \boldsymbol{t}^0_j \rangle_2 + \frac{\left( \cos_\kappa \left(\sqrt{\kappa}\right) \right)^2}{ \kappa}  \\
% &= \frac{\left( \sin_\kappa \left(\sqrt{\kappa}\right) \right)^2}{ \kappa}\mathcal K^{E}(t_i,t_j) + \frac{\left( \cos_\kappa \left(\sqrt{\kappa}\right) \right)^2}{ \kappa} \\
% &= \psi^R(t_i - t_j),
% \end{aligned}
% \end{equation}
That is, for arbitrary $\kappa$, we have $\mathcal K^{R}(t_i,t_j)=\psi^R(t_i - t_j)$ and $\psi^R=g \circ \psi^E$, where $g(x)=Ax+B$,
\vspace{-0.07in}
\begin{equation}
A=\frac{\left( \sin_\kappa \left(\sqrt{|\kappa|}\right) \right)^2}{ |\kappa|},  \ \ B=\frac{\left( \cos_\kappa \left(\sqrt{|\kappa|}\right) \right)^2}{ |\kappa|}.
\vspace{-0.05in}
\end{equation}
\end{proof}

\vspace{-0.2in}
\subsubsection{Riemannian Temporal Attention Layer} 

We propose a Riemannian temporal attention layer, which is the building block layer of $\mathcal{R}$\textsc{GNN} to update temporal representations $\boldsymbol h_i(t)$ in the general Riemannian space. 
As opposed to static graph convolution receiving messages of all members in the neighborhood, we conduct temporal aggregation on the \emph{temporal neighbors} (\textsc{Definition 2}) $\mathcal N_t(v_i)$ at time $t$ in the following two steps.

\noindent{\textbf{Step One.} \emph{We build the temporal message with the stereographic projection in the Riemannian manifold.}
For a neighbor node $v_j$ linked at time $t_l$, its temporal message is built by the representation of nodes $v_j$ and time  encoding $\varphi^\kappa(t_l)$ in the Riemannian manifold. 
However, operating vectors in the  Riemannian manifold is nontrivial, and concatenation is generally illegal in the Riemannian manifold.
Fortunately,  addition is well-defined in the gyrovector spaces  $\oplus_{G}$ with the elegant non-associative algebraic formalism:
\vspace{-0.1in}
\begin{equation}
\boldsymbol{x} \oplus_{G} \boldsymbol{y}=\frac{\left(1-2 \kappa \boldsymbol{x}^{T} \boldsymbol{y}-\kappa\|\boldsymbol{y}\|^{2}\right) \boldsymbol{x}+\left(1+\kappa\|\boldsymbol{x}\|^{2}\right) \boldsymbol{y}}{1-2 \kappa \boldsymbol{x}^{T} \boldsymbol{y}+\kappa^{2}\|\boldsymbol{x}\|^{2}\|\boldsymbol{y}\|^{2}}.
\vspace{-0.03in}
\end{equation}
The mappings between the gyrovector space and the Riemannian manifold of Eq. (\ref{manifold}) is done via the stereographic projection $SP(\cdot)$ and its inverse as follows:
% \begin{equation}
% \begin{aligned}
% SP(\boldsymbol{x}) & =\frac{1}{1+\sqrt{\kappa} \boldsymbol{x}_{[1]}} \boldsymbol{x}_{[2: d+1]} \\
% SP^{-1}(\boldsymbol x^\prime) & =\left[    \frac{1}{\sqrt{\kappa}}    \left(\lambda_{\boldsymbol x^\prime}^{\kappa} -1 \right), \lambda_{\boldsymbol x}^{\kappa} \boldsymbol x^\prime  \right],
% \end{aligned}
% \end{equation}
\vspace{-0.035in}
\begin{equation}
SP(\boldsymbol{x})  =\frac{1}{1+\sqrt{\kappa} \boldsymbol{x}_{[1]}} \boldsymbol{x}_{[2: d+1]}, \
SP^{-1}(\boldsymbol x^\prime)  =\left[    \frac{1}{\sqrt{\kappa}}    \left(\lambda_{\boldsymbol x^\prime}^{\kappa} -1 \right), \lambda_{\boldsymbol x}^{\kappa} \boldsymbol x^\prime  \right],
\end{equation}
where $\lambda_{\boldsymbol{x}}^{\kappa}=\frac{2}{ 1+\kappa \| \boldsymbol{x} \|_2^2 }$, and $\boldsymbol x^\prime$ is the corresponding point of $\boldsymbol x$ in the gyrovector space. Thus, we have the addition in the general Riemannian manifold  as follows:
\vspace{-0.035in}
\begin{equation}
\boldsymbol x \oplus_{\kappa} \boldsymbol y=SP^{-1}\left(SP\left(\boldsymbol x\right) \oplus_{G} SP\left(\boldsymbol y\right)\right),
\end{equation}
and formulate the temporal message from $v_j$ as follows:
\begin{equation}
\boldsymbol m_{j}(t) =(\boldsymbol W_1 \otimes_{\kappa} \boldsymbol h_j(t)) \oplus_{\kappa} (\boldsymbol W_2 \otimes_{\kappa}\varphi^\kappa\left(t_l\right)),
\end{equation}
%$\boldsymbol m_j(t) =(\boldsymbol W_1 \otimes_{\kappa} \boldsymbol h_j(t)) \oplus_{\kappa} (\boldsymbol W_2 \otimes_{\kappa}\boldsymbol t^\kappa_j)$, 
\noindent where $\boldsymbol W_1$ and $\boldsymbol W_2$ are parameter matrices.
%and note that $\otimes_{\kappa}$ has a higher priority than $\oplus_{\kappa}$. 
Owing to \emph{the translation invariant with respect to time}, we alternatively set $\bar t_l=t-t_l$ in practice as $\mathcal K^R(\bar t_i, \bar t_2)=\mathcal K^R(t_1, t_2)$ for any $t_1, t_2$ in the time domain.

\noindent{\textbf{Step Two.} \emph{We perform the aggregation with the attention mechanism in the Riemannian space.}
As the importance of neighbor nodes is usually different, we introduce the attentional aggregation in account of the importance of different neighbors.
Specifically, we first lift the Riemannian temporal message to the tangent space via logarithmic map $log _\mathcal O^{\kappa}$ and model the importance parameterized by $\boldsymbol{\theta}$  and $\mathbf W_3$  as follows:
\vspace{-0.035in}
% \begin{equation}
% ATT^{\kappa}(\boldsymbol m_{ij}(t))= exp_\mathcal O^{\kappa} \left( \sigma \left(  log_\mathcal O^{\kappa}(\boldsymbol W_3 \otimes_{\kappa} \boldsymbol m_{ij}(t)  \right) \right),
% \end{equation}
\begin{equation}
ATT^{\kappa}(\boldsymbol m_i(t), \boldsymbol m_j(t))= \sigma \left( \boldsymbol{\theta}^\top \left[ \boldsymbol W_3 log _\mathcal O^{\kappa}(\boldsymbol m_i(t)) , \boldsymbol W_3 log_\mathcal O^{\kappa}(\boldsymbol m_j(t)) \right] \right),
\end{equation}
where $\sigma(\cdot)$ denotes the sigmoid activation. 
Then, we compute the attention weight over the temporal neighborhood via softmax:
\begin{equation}
\alpha_{ij}=e^{  ATT^{\kappa}( \boldsymbol m_{i}(t), \boldsymbol m_{j}(t)   )}  / \sum\nolimits_{v_n \in \mathcal N_i(t)}e^{  ATT^{\kappa}( \boldsymbol m_{i}(t),\boldsymbol m_{n}(t)  )}.
\end{equation}
Finally, we update $\boldsymbol h_i(t)$ by performing the attentional aggregation with the aid of the tangent space, i.e., 
\begin{equation}
AGG^{\kappa}\left(\{\boldsymbol m_j(t)\}\right) =\delta^{\otimes_\kappa}\left(exp _{\boldsymbol h_i(t)}^{\kappa}\left(\sum\nolimits_{j \in \mathcal{N}_t(v_i)} \alpha_{i j} \ log_{\boldsymbol h_i(t)}^{\kappa}\left(\boldsymbol m_j(t)\right)\right)\right),
\end{equation}
where $\delta$ is the applied nonlinearity in the Riemannian space.

%Next,  we formulate the parametric curvature in the temporal attention.
\subsubsection{Functional Curvature in General Riemannian Space} 
% We first present the temporal attention , and then formulate the parameteric expression of the curvature to model the structure evolvement of the temporal graph over time.
% \noindent\emph{\textbf{Parameteric Curvature}}: 
We propose a novel functional curvature over time, a key ingredient of our $\mathcal{R}$\textsc{GNN}.
%The curvature is to describe the shape of embedding Riemannian space of a graph. 
% Hence, we propose a parametric curvature as the function of time to arbitrary curvature of the general Riemannian space, so that we are able to match the underlying geometry of the temporal graph at any time, either hyperbolic ($\kappa <0$) or hyperspherical ($\kappa >0$),  without restriction given in prior.
In this way, we can model the evolvement shifting among the Riemannian space of various curvatures over time, and distinguishes us against the studies restricting in a certain curvature with the inductive bias given in prior.

In the Riemannian geometry, the evolvement of temporal graph is naturally described as the time-varying curvature of the embedding space.
%the structure evolvement of temporal graph  is naturally described as the curvature of its embedding space evolves over time. 
Specifically, we need to figure out a curvature function over time $f: t \to \kappa$. In practice, we first perform $\varphi^0(t)$ to obtain a time encoding $\boldsymbol t^0$, and then feed the encoding vector into a neural network, CurNN. The CurNN is built with an MLP followed by a bilinear output layer to obtain the constant curvature of any sign, 
%i.e., we have the function as follows:
\vspace{-0.05in}
\begin{equation}
CurNN(t)=MLP\left(\varphi^0(t)\right)^\top \boldsymbol W_4 MLP\left(\varphi^0(t)\right).
\vspace{-0.05in}
\label{cur}
\end{equation}
%\noindent where $\boldsymbol W_4$ is the weight matrix of the bilinear output layer.
With the functional curvature in Eq. (\ref{cur}), $\mathcal{R}$\textsc{GNN} is able model the graph evolvement shifting among hyperspherical, Euclidean and  hyperbolic spaces over time.

%where we adopt the time encoding $\boldsymbol t^0$ as input. 

\vspace{-0.05in}
\subsection{Riemannian Self-supervised Learning}
\vspace{-0.02in}
To enable the self-supervised learning, we propose a novel Riemannian self-supervised learning approach. The novelty lies in that we explore the rich information in the Riemannian space of the temporal graph itself, getting rid of the effort for data augmentation.
The self-supervised learning  tasks are dual, i.e., the temporal representation and the curvature of the Riemannian Space.
To this end, we propose a Riemannian reweighted self-contrastive learning and an edge-based self-supervised curvature learning for the temporal graph, respectively.

\vspace{-0.05in}
\subsubsection{Reweighted Self-Contrastive Learning in the Riemannian Space} 
Contrastive learning explores the semantic similarity of data themselves, and learn the representations by contrasting positive and negative samples \cite{HardNeg,DDGCL}.
Consider that a latent semantic class $c \in \mathcal C$ is assigned to each observation $x$ over $\mathcal X$ via $h: \mathcal X \to \mathcal C$.
Given an observation $x$, if  $x^\prime$ and $x$ share the same semantic class, $x^\prime$ is said to be a positive sample whose conditional distribution is given as $p^{+}\left(x^{\prime}\right)=p\left(x^{\prime} \mid h\left(x^{\prime}\right)=h(x)\right)$, while a negative sample is drawn from $p^{-}\left(x^{\prime}\right)=p\left(x^{\prime} \mid h\left(x^{\prime}\right) \ne h(x)\right)$.
We cannot access to the sampling distributions $p^+$ and $p^-$  in practice.
The main ingredients of a contrastive learning framework are: 
i) proxies of $p^+$ of a given node $v_i$, and ii) a loss function discriminating positive and negative samples.
Unfortunately, both of them are challenging in the context of temporal graphs.

\noindent\textbf{Riemannian Self-augmentation.}
For the first challenge (obtaining $p^+$), data augmentation is usually performed and thereby different views are constructed for contrast.
In computer vision, the augmentation can be easily given by semantic preserving transformations, e.g.,  cropping and rotating \cite{ChenK0H20}. However, the analog is not obvious for graphs, and the study \cite{HassaniA20} shows that different augmentations (i.e., node dropping, edge perturbation) behave differently according to the distributions of the underlying graphs.

To address this challenge, we propose a novel Riemannian self-augmentation, which leverages the functional curvature to augment auxiliary views.
%information of temporal graph.
In this way, we obtain $p^+$ without introducing a new graph, instead of struggling in defining augmented graphs as prior works \cite{VelickovicFHLBH19,HassaniA20,QiuCDZYDWT20}.
%we present a self-augmentation benefited from the merit of Riemannian geometry, which leverages the information of temporal graph itself .
% evolvement of temporal graphs. 
Specifically, given the $\alpha$ view at time $t_1$, the proposed self-augmentation aims to generate its $\beta$ view for the contrastive learning.
%for contrast without introducing a new graph.
The temporal representations of $\alpha$ view are obtained via $\mathcal{R}$\textsc{GNN}.
%with the unified formalism of Riemannian manifold, 
Alternatively, we can employ another time point $t_2$ as a reference and infer the temporal representations of  $\beta$ view at $t_1$ based on corresponding curvatures.
Thanks to \emph{the proposed functional curvature over time}, we can obtain the curvatures in the evolvement of temporal graph via Eq. (\ref{cur}), and generate the $\beta$ view with a Riemannian projection as follows:
\vspace{-0.1in}
\begin{equation}
RiemannianProj_{t_2 \to t_1}(\cdot)=exp^{CurNN(t_1)}_\mathcal O \left(log^{CurNN(t_2)}_\mathcal O \left( \cdot \right)\right).
\vspace{-0.05in}
\end{equation}
%\noindent where the curvatures  are defined via Eq. (\ref{cur}).

\noindent\textbf{Riemannian Reweighting.}
For the second challenge (constrastive loss), different formulations of constrastive loss are proposed. 
However, in practice, the ideal negative sampling distribution $p^-$ is replaced by the data distribution $p(x)$ over $\mathcal X$ (\emph{sampling bias}), since labels cannot be accessed in the self-supervised learning. Additionally, the ``negative'' samples behave uniformly in the constrastive objective (\emph{hardness unawareness}). 
The phenomena are formalized in \citet{HardNeg}, however, its solution cannot be applied in the Riemannian space owing to the essential distinction in geometry.

%incorporate the data distribution $p(x)$ with positive samples to
To bridge this gap, we propose a novel Riemannian reweighting contrastive loss to i) get rid of the sampling bias as well as ii) select negative samples in account of the hardness. Specifically, we first confront the bias incurred by the absence of ideal $p^-$ with tractable distributions. We decompose the data distribution as $p\left(x^{\prime}\right)=\tau^{+} p^{+}\left(x^{\prime}\right)+\tau^{-} p^{-}\left(x^{\prime}\right)$, where $\tau^+ =p(h(x^\prime)=h(x))$ is the class prior of $x$’s semantic class and can be estimated from data in practice \cite{JainWR16}, and thus $p^-$ is yielded as
\vspace{-0.03in}
\begin{equation}
p^{-}\left(x^{\prime}\right)=\left(p\left(x^{\prime}\right)-\tau^{+} p^{+}\left(x^{\prime}\right)\right) / \tau^{-},
\vspace{-0.05in}
\end{equation}
with two tractable distributions. Note that we have samples from $p$ and $p^{+}$ is given via the self-augmentation above.
Second, we introduce a probability $q^-_\xi$ in Riemannian space to select negative samples. A hard negative sample is an $x^-$ whose semantic class is  different from the $x$ but the representation is similar to $x$, and thus $q_{\xi}^{-}\left(x^{-}\right)=q_{\xi}\left(x^{-} \mid h(x) \neq h\left(x^{-}\right)\right)$ is defined as 
\vspace{-0.05in}
\begin{equation}
q_{\xi}\left(x^{-}\right) \propto e^{\xi s_\mathcal M(x, x^-)} \cdot p\left(x^{-}\right),
\vspace{-0.03in}
\label{reweight}
\end{equation}
\noindent where $s_\mathcal M$ is a score function to output the similarity, and $\xi >0$ is the parameter to upweight the hardness.
The intuition is that a hard negative with similar representation in Riemannian space has a larger probability of getting sampled.
With the prior of $p$, the negative sampling is \emph{essentially reweighted by the likelihood of an exponential term}, which is the Bayesian interpretation of Eq. (\ref{reweight}).

Obviously, it further requires a score function to contrast between positive and negative samples.
Defining the score function is nontrivial as the existing Euclidean functions cannot be used in the Riemannian space, and temporal information needs to be considered for representation learning on temporal graphs.
Thanks to the \emph{translation invariant property} of the proposed time encoding of arbitrary curvature, we propose a novel score function as follows:
\begin{equation}
s_\mathcal M\left(\boldsymbol h_i(t), \boldsymbol h_i(t)\right)= \mathcal K(t_i,t_j) d_\mathcal M\left(\boldsymbol h_i(t), \boldsymbol h_i(t)\right),
\label{sim}
\vspace{-0.03in}
\end{equation}
\noindent which means that the samples are discriminated by the distance in the manifold penalized by the relative relationship in time domain.

\begin{algorithm}[tb]
        \caption{The Self-supervised Learning of \textsc{Self}$\mathcal{R}$\textsc{GNN}} 
        % \LinesNumbered
        \KwIn{A temporal graph $\mathcal G = (\mathcal V, \mathcal E, \boldsymbol{X}, T)$, weighting coefficient $w$.}
        \KwOut{The parameters of \textsc{Self}$\mathcal{R}$\textsc{GNN}.}
        \While{not converging}{
            %Draw a congruent graph generation function $\Gamma(\cdot)$\;
            \textcolor{cyan}{\emph{//  The $\alpha$ view:}}\\
            Estimate curvature at the time $t_1$ via Eqs. (\ref{EucEncoding}) and (\ref{cur});\\
            $\boldsymbol h(t_1) \gets CurvatureVarying\mathcal{R}GNN$  at $t_1$;\\
            $\boldsymbol h(t)_\alpha  \gets \boldsymbol h(t_1)$ ;\\
            \textcolor{cyan}{\emph{//  The $\beta$ view from Riemannian self-augmentation:}}\\
            Estimate curvature at the time $t_2$ via Eqs. (\ref{EucEncoding}) and (\ref{cur});\\
            $\boldsymbol h(t_2) \gets CurvatureVarying\mathcal{R}GNN$  at $t_2$;\\
            $\boldsymbol h(t)_\beta  \gets RiemannianProj_{t_2 \to t_1}\left(\boldsymbol h(t_2)\right)$ ;\\
            \textcolor{cyan}{\emph{//  Riemannian self-supervised learning objective:}}\\
             \For{temporal representations in $\alpha$ and $\beta$ views}{
                      Contrast via the score function in  Eq. (\ref{sim});\\
                      Calculate Riemannian reweighting contrastive loss $\mathcal L_{(\alpha, \beta)}$ and $\mathcal L_{(\beta, \alpha)}$ via Eqs. (\ref{obj1})-(\ref{obj3});\\
            }
             $\hat \kappa  \gets GRU([\kappa_{ij} || \boldsymbol t_j^0])$ ;\\
             \textcolor{cyan}{\emph{// Update neural network parameters:}}\\
              Calculate the gradients of the overall objective: \\
              \vspace{-0.1in}
            $$ \nabla \left( \mathcal L_{contrast} + w \mathcal L_{curvature} \right). $$
           %  Set $\mathbf G^\alpha = \mathbf G$\;
           %  $\mathbf Z^\alpha = \emph{MixedCurvatureGNN}(\mathbf G^\alpha, \mathbf X; \boldsymbol{\theta}^\alpha)$\;
           %  $[\mathbf H^\alpha \ \mathbf E^\alpha \ \mathbf S^\alpha] = \emph{RiemannianProjector}(\mathbf Z^\alpha; \boldsymbol{\phi})$\;
           %  \textcolor{cyan}{\emph{//  Views in the congruent augmentation $\mathbf G^\beta$:}}

           % Generate a congruent graph  $\mathbf G^\beta = \Gamma(\mathbf G)$\;
           %  $\mathbf Z^\beta = \emph{MixedCurvatureGNN}(\mathbf G^\beta, \mathbf X; \boldsymbol{\theta}^\beta)$\;
           %  $[\mathbf H^\beta \ \mathbf E^\beta \ \mathbf S^\beta] = \emph{RiemannianProjector}(\mathbf Z^\beta; \boldsymbol{\phi})$\;
           %  \textcolor{cyan}{\emph{// Dual contrastive loss:}}

           %  \For{each node $v_i$ in $\mathbf G^\alpha$ and $v_j$ to $\mathbf G^\beta$}{
           %          \For{Riemannian views $\mathbf x,  \mathbf y \in \{\mathbf h, \mathbf e, \mathbf s\}$}{
           %                  %Compute Eqs. (\ref{alpha_1}) and (\ref{alpha_2}) with the discriminator $\mathcal D^\kappa(\mathbf x^\alpha, \mathbf x^\beta; \mathbf D_T)$\;
           %                  Single-view contrastive learning with Eqs. (\ref{alpha_1}) and (\ref{alpha_2})\; 
           %                  Cross-view contrastive learning with Eqs. (\ref{geo_1}) and (\ref{geo_2})\; 
           %          }
           %  }
           %  \textcolor{cyan}{\emph{// Update neural network parameters:}}

           %  Compute gradients of the dual contrastive loss:
           %     \vspace{-0.05in}
           %  $$
           %  \nabla_{\boldsymbol{\theta}^\alpha, \ \boldsymbol{\theta}^\beta, \ \boldsymbol{\phi},\ \mathbf D_S, \ \mathbf D_C}\ \ \mathcal J_S + \lambda \mathcal J_C. 
           %  $$
           %     \vspace{-0.2in}
            }
            \label{algo}
\end{algorithm}

\subsubsection{Edge-based Self-supervised Curvature Learning}
%In \textsc{Self}$\mathcal{R}$\textsc{GNN}, we leverage Ricci curvature to boost the curvature learning of the temporal graph.
We propose to utilize the Ricci curvature on the edges to supervise the functional curvature of the graph.
The (coarse) Ricci curvature $\kappa_{ij}$ on edge $(v_i, v_j)$  is defined by comparing the Wasserstein distance $W(m^\lambda_i, m^\lambda_j)$ to the geodesic distance $d_\mathcal M(\boldsymbol h_i(t), \boldsymbol h_j(t))$ on the manifold \cite{RcciOlli}, 
%where $m^\alpha_i$ is a probability measure of total mass $1$ around an end node $v_i$ of the edge, i.e., 
where $m^\alpha_i$ is a probability measure around node $v_i$, i.e.,
\vspace{-0.09in}
\begin{equation}
\kappa_{ij}=1- W(m^\lambda_i, m^\lambda_j)/d_\mathcal M(\boldsymbol h_i(t), \boldsymbol h_j(t)).
\vspace{-0.03in}
\end{equation}
Given a node $v_i$ with the temporal neighborhood $\mathcal N_{t}(v_i)$, the probability measure $m_{i}^{\lambda}$ is defined as 
\vspace{-0.07in}
\begin{equation}
m_i^{\lambda}\left(v \right)= \begin{cases}\lambda & \text { if } v=v_i \\ (1-\lambda) / K & \text { if } v \in \mathcal N_{t}(v_i), \end{cases}
\vspace{-0.05in}
\end{equation}
% \begin{equation}
% m_i^{\lambda}\left(v \right)= \{\lambda & \text { if } v=v_i,  \ (1-\lambda) / K & \text { if } v \in \mathcal N_{t}(v_i),  \ 0 & \text { otherwise} \},
% \end{equation}
\noindent and otherwise, $m_i^{\lambda}\left(v \right)=0$, where $K$ is the number of the nodes in the neighborhood, and $\alpha$ is the parameter to keep probability mass of $\alpha$ at node $v_i$ itself, which is set to $0.5$ in practice according to the study of \citet{YeLM0020}.
For a given time point $t$, the curvature of the graph is induced from the Ricci curvatures \cite{1997Riemannian}, and we employ a GRU to mimic the mapping. 
Concretely, we first pair the Ricci curvature of an edge $\kappa_{ij}$ with its time encoding $\boldsymbol t_j^0$ to incorporate the temporal information. Then, we feed the augmented $[\kappa_{ij}||\boldsymbol t_j^0]$ into the GRU unit in chronological order, whose output layer is replaced by a bilinear one to obtain the graph curvature $\hat \kappa$ of any sign.
$\hat \kappa$ is presented as the supervision signal to the graph curvature $\kappa$ given by function of $CurNN(t)$, and thus we have the objective of  self-supervised curvature learning, i.e., $\mathcal L_{curvature}=\sum\nolimits_t|\kappa-\hat \kappa|$.

%\begin{aligned}x  & \sim p \\ x^+  & \sim p_x^+ \end{aligned} 
\subsubsection{Self-supervised Learning Objective}
First, we instantiate the formulation of Riemannian reweighted self-contrastive learning. We start with defining $\mathcal L_{(\alpha, \beta)}$ that contrasts with $\boldsymbol h(t)_\beta^+ \sim p^+$ and $\{\boldsymbol h_i(t)_\beta^-\}_{i=1}^N \sim p^-$ from the self-augmented $\beta$ view as follows:
\vspace{-0.07in}
\begin{equation}
\vspace{-0.05in}
\mathbb{E}_{\boldsymbol h(t)_\alpha, \boldsymbol h(t)_\beta^+}\left[s_\mathcal M(\boldsymbol h(t)_\alpha, \boldsymbol h(t)_\beta^+)-\mathbb{E}_{\boldsymbol h_i(t)_\beta^-}\left[    log\sum_{i=1}^{N} e^{\left(s_\mathcal M(\boldsymbol h(t)_\alpha, \boldsymbol h_i(t)_\beta^-\right)}  \right]\right],
\label{obj1}
\end{equation}
and the expectation term of $\mathbb{E}_{\boldsymbol h(t)_\beta^-}$ is replaced by the Riemannian reweighting accordingly,
\vspace{-0.07in}
\begin{equation}
\vspace{-0.05in}
\frac{1}{\tau^{-}Z_{\xi}}\left(\mathbb{E}_{\boldsymbol h_i(t)_\beta^-}\left[ e^{s_\mathcal M(\boldsymbol h(t)_\alpha, \boldsymbol h_i(t)_\beta^-)} \right]- \frac{\tau^{+}}{Z_{\xi}^-} \mathbb{E}_{\boldsymbol h(t)_\beta^+}\left[ e^{s_\mathcal M(\boldsymbol h(t)_\alpha, \boldsymbol h(t)_\beta^+)} \right]\right),
\label{obj2}
\end{equation}
\noindent where $q_{\xi}^{+} (\boldsymbol h(t)_\beta^-)\propto e^{\xi s_\mathcal M(\boldsymbol h(t)_\alpha, \boldsymbol h(t)_\beta^+))} \cdot p^{+}(\boldsymbol h(t)_\beta^-)$, and the factors 
\vspace{-0.07in}
\begin{equation}
Z_{\xi}=\mathbb E_{\boldsymbol h_i(t)_\beta^-}\left[e^{\xi s_\mathcal M(\boldsymbol h(t)_\alpha, \boldsymbol h_i(t)_\beta^-)}\right],  \  Z_{\xi}^{+}=\mathbb E_{\boldsymbol h(t)_\beta^+}\left[e^{\xi s_\mathcal M(\boldsymbol h(t)_\alpha, \boldsymbol h(t)_\beta^+)}\right],
\label{obj3}
\end{equation}
\noindent are given to normalize the probability mass. Note that, if we set hardness $\xi=0$, the formulation in Eqs. (\ref{obj1})-(\ref{obj3}) degenerates into a typical InfoNCE loss treating the samples uniformly, and in fact it is easy to check the following proposition holds.
\newtheorem*{prop3}{Proposition 3 (Riemannian Reweighting Contrastive Loss)} 
\begin{prop3}
The  proposed  formulation  in Eqs. (\ref{obj1})-(\ref{obj3})  is equivalent to InfoNCE objective formulated in \cite{DeepInfoMax} if class prior is omitted, $\tau^+=0$.
\end{prop3}
\noindent That is, \emph{we generalize the formulation of InfoNCE in the Riemannian space with reweighting}. The $\alpha$ view is contrasted with the $\beta$ view, and vice versa. We have $\mathcal L_{contrast}=-\mathcal L_{(\alpha, \beta)}-\mathcal L_{(\beta, \alpha)}$.

Finally, incorporating the objective of  curvature learning, we have the overall objective for Riemannian self-supervised learning,
  \vspace{-0.1in}
\begin{equation}
\mathcal L_{self}=\mathcal L_{contrast} + w \mathcal L_{curvature},
\end{equation}
\noindent where $w$ is a weighting coefficient.
We summarize the self-supervised learning of the proposed \textsc{Self}$\mathcal{R}$\textsc{GNN} in Algorithm \ref{algo},
%\textsc{Self}$\mathcal{R}$\textsc{GNN} learns the temporal representations in the Riemannian space of arbitrary curvature in absence of external guidance.
whose computational complexity is $O(N_e(|\mathcal E|+N_w|\mathcal V|))$, where $N_e$ and $N_w$ are the numbers of epochs and reweighted samples, respectively. 
Note that, the complexity order of  \textsc{Self}$\mathcal{R}$\textsc{GNN} is same as that of self-supervised graph methods in Euclidean space \cite{DDGCL,HassaniA20}, but \emph{we  support time-varying curvature to model the evolvement shifting among hyperspherical, Euclidean and hyperbolic spaces over time}.

  \vspace{-0.02in}
\section{Experiment}
In this section, we conduct extensive experiments on a variety of datasets, aiming to answer the following research questions (\emph{RQs}):
\begin{itemize}
  \item \textbf{\emph{RQ1}}: How does the proposed \textsc{Self}$\mathcal{R}$\textsc{GNN} perform?

  \item \textbf{\emph{RQ2}}: How does each component contributes to the success of the proposed \textsc{Self}$\mathcal{R}$\textsc{GNN}?

    \item \textbf{\emph{RQ3}}: How does the curvature evolve over time?

\end{itemize}

 \begin{table}
    \centering
          \caption{The summary of AUC (\%) for node classification on Wikipedia, MOOC and Cora datasets. The highest scores are in \textbf{bold}, and the second \emph{\underline{underlined}}.}
          %\vspace{-0.1in}
    \begin{tabular}{c c l | c c c}
      \toprule
  \multicolumn{3}{c|}{\textbf{Model}} &  \textbf{Wiki} & \textbf{MOOC} & \textbf{Cora}\\
     \toprule
     \multirow{7}{*}{\rotatebox{90}{\textbf{Supervised}} } 
       &\multirow{5}{*}{ $\mathbb E$ } 
                         &EvolveGCN   
                          & $  72.33(0.6)$     & $65.35(0.1)$     &$  76.10(0.3)$ \\
             &          &  VGRNN  
                          &  $ 80.15(0.1)$   &  $ 71.02(0.2)$   &  $  82.05(0.2)$   \\
              &          &  DyRep  
                          &  $ 79.24(0.2)$   &  $72.67(0.0)$     & $  82.89(0.1)$    \\
              &             & TGAT  
                        &  $ 83.69(0.7)$   &  $ 69.46(0.4)$    & $  85.27(0.2)$   \\
              &         &  CAWNet 
                         & $  86.77(0.3)$   &  $ 68.77(0.4)$    &$  88.95(0.7)$   \\
                         \cline{2-6}
         &\multirow{2}{*}{$\mathbb H$}
                        &  HVGNN  
                        &  $ 86.22(0.2)$   &  $  73.90(0.3)$   &  $ \emph{\underline{89.48}}\ (0.1)$   \\
            &              &  HTGN  
                          &  $ 85.08(0.5)$   &  $ \emph{\underline{75.12}}\  (0.1)$   &  $ 87.22(1.0)$   \\
      \midrule 
       \multirow{3}{*}{\rotatebox{90}{\textbf{Self}}} 
                     & $\mathbb E$    &   DDGCL 
                        & $  \emph{\underline{89.32}}\ (0.5)$     & $74.54(0.2)$     &$  87.67(0.3)$ \\
                        \cline{2-6}
    &  &  \textbf{\textsc{Self}$\mathcal{R}$\textsc{GNN}}
                        &  $ \mathbf{93.64}(0.\  )$   &  $  \mathbf{81.28}(0.1)$   &  $ \mathbf{94.06}(0.2)$   \\
    &  &  (Gain)
                        &  $+4.32\% $   &  $ +6.16\% $   &  $+5.58\% $   \\
      \bottomrule
    \end{tabular} 
        \label{results1}
                  \vspace{-0.05in}
  \end{table}

   \begin{table*}
    \centering
          \caption{The summary of AUC($\%$) for \emph{inductive link prediction} and \emph{transductive link prediction} on Wiki, MOOC, Cora, Social and Physics. The highest scores are in \textbf{bold}, and the second \emph{\underline{underlined}}. Note, (0. ) means that the interval is less than $\pm0.05\%$.}
                            \vspace{-0.1in}
       \resizebox{1.02\linewidth}{!}{ 
  \begin{tabular}{p{0.2cm}<{\centering} p{0.15cm}<{\centering} p{1.6cm} | p{1.22cm}<{\centering} p{1.22cm}<{\centering} p{1.22cm}<{\centering} p{1.22cm}<{\centering} p{1.22cm}<{\centering} | p{1.22cm}<{\centering}  p{1.22cm}<{\centering} p{1.22cm}<{\centering}  p{1.22cm}<{\centering} p{1.22cm}<{\centering}}
      \toprule
  \multicolumn{3}{c|}{}        & \multicolumn{5}{c|}{ \emph{Inductive Link Prediction} } &  \multicolumn{5}{c}{ \emph{Transductive Link Prediction} } \\
  \multicolumn{3}{c|}{ \textbf{Method} }  & \textbf{Wiki} & \textbf{MOOC}  & \textbf{Social} & \textbf{Cora} & \textbf{Physics} 
                                                                      & \textbf{Wiki} & \textbf{MOOC} & \textbf{Social} & \textbf{Cora} & \textbf{Physics}\\
     \toprule
          \multirow{7}{*}{\rotatebox{90}{\textbf{Supervised}} } 
      & \multirow{5}{*}{ $\mathbb E$}
                         &EvolveGCN    
&  $ 57.26(1.2)$   &  $  51.52(2.4)$   &  $ 48.85(1.0)$  & $  69.18(0.3)$   & $ 74.25(1.6)$  
                                                                                      &  $ 60.48(0.5)$    &  $50.36(0.8)$  & $60.36(0.6)$ &  $ 68.02(0.1)$ &  $ 77.50(1.3)$ \\
       &               & VGRNN 
&  $ 62.40(0.7)$   & $  61.33(0.5)$   &  $  65.48(0.8)$  & $  75.67(0.1)$    &$ 73.88(1.0)$    
                                                                                       &  $71.20(0.7)$   &  $90.02(0.3)$  & $78.28(0.7)$ &  $ 79.55(0.2)$ &  $78.02(1.0)$ \\
        &               & DyRep 
&  $ 73.39(1.0)$    &  $84.23(1.8)$     & $ 86.44(0.2)$ &  $  81.30(0.1)$    & $76.05(1.3)$    
                                                                                       &  $ 77.40(0.1)$   &  $90.49(0.\ )$  & $90.85(0.\ )$ &   $82.13(0.2)$ & $78.67 (0.7)$\\
         &                & TGAT  
&  $ 95.20(0.6)$    &  $ 69.33(0.1)$     & $ 53.79(1.1)$  & $  76.92(0.2)$    & $81.46(0.2)$    
                                                                                       &  $ 96.36(0.1)$   &  $72.09(0.3)$   & $56.63(0.5)$ &   $77.36(0.1)$ &   $ 95.12(1.1)$ \\
      &              & CAWNet 
&  $ \emph{\underline{98.24}}\ (0.5)$     &  $90.67(0.6)$      & $ 95.15(0.7)$ &  $  \emph{\underline{95.20}}\ (0.5)$    & $95.12(0.1)$    
                                                                                    &  $ \mathbf{99.89}(0. \ )$   &  $92.38(0.6)$  & $94.79(0.2)$  &  $95.89(0.3)$  &  $\emph{\underline{97.86}}\ (0.7)$\\
 \cline{2-13}
     & \multirow{2}{*}{$\mathbb H$}
                      &  HVGNN 
&  $ 96.55(0.4)$   &  $95.20(0.\ )$     & $ 89.12(0.1)$ &  $  93.04(1.0)$    & $\emph{\underline{96.02}}\ (0.3)$    
                                                                                         &  $ 98.62(0.1)$   &  $89.33(0.2)$  & $84.67(0.1)$ &  $93.67(0.5)$  &  $ 97.33(0.1)$\\
        &                  &  HTGN 
&  $ 87.17(0.5)$    & $91.48(0.6)$     & $\emph{\underline{97.33}}\ (0.2)$   &   $ 95.15(0.7)$ &  $ 89.24(0.2) $     
                                                               &  $ 91.75(0.7)$   &    $ 95.10(1.1)$ & $\emph{\underline{98.05}}\ (1.0)$  & $\emph{\underline{96.27}}\ (0.2)$    &   $91.67(0.1)$ \\
     \midrule 
      \multirow{2}{*}{\rotatebox{90}{\textbf{Self} }} 
   &  $\mathbb E$   &   DDGCL 
&  $ 98.05(0.1)$     &  $\emph{\underline{96.16}}\ (0.7)$      & $ 97.08(0.4)$ &  $  94.89(0.3)$    & $95.70(0.2)$    
                                                                                       &  $ 97.92(0.1)$    &  $\emph{\underline{96.92}}\ (0.1)$   & $95.18(0.1)$ &  $95.05(0.6)$  &  $ 96.10(0.3)$\\
       \cline{2-13}
  &  & \textbf{\textsc{Self}$\mathcal R$\textsc{GNN}}
  &  $\mathbf{99.16}(0.2)$  & $\mathbf{98.12}(0.1)$   & $\mathbf{97.80}(0.1)$   &   $\mathbf{96.36}(0.3)$ &  $\mathbf{97.99}(0.1)$     
                                &  $\emph{\underline{99.67}}\ (0.1)$   &   $\mathbf{98.85}(0.1)$   & $\mathbf{99.24}(0.\ )$  & $\mathbf{97.68}(1.0)$    &   $ \mathbf{98.05}(0.2)$ \\
      \bottomrule
    \end{tabular} }
 %   }
 \vspace{-0.05in}
        \label{results2}
  \end{table*}
  
            \vspace{-0.1in}
\subsection{Experimental Setups}

\subsubsection{Datasets} 
We conduct extensive experiments on a diverse set of benchmark temporal graphs including \textbf{Wikipedia}, \textbf{MOOC} and \textbf{Social} of \citet{Xu2020inductive}, 
and \textbf{Cora} and \textbf{Physics} of \citet{2019Variational}.
% Specifically, 
% Wikipedia is a graph constructed by the temporal interactions between wiki pages and human editors, where an interaction represents a user editing a page. 
% MOOC is a temporal graph of students and online course units with the interactions describing the user actions on the courses.
% Cora is originally a citation network where the nodes in the graph represent the publications and the edges indicate the citation relations, and links are then timestamped by the study \cite{2019Variational}.
% Social is a temporal friendship network that records evolving physical proximity between students over a year, determined by wireless signals of their mobile devices.
% Physics is a citation network related to high energy physics, collected from the e-print arXiv website, and the temporal information is given by the submission time.
We adopt the same chronological data split with $70\%$ for training, and $15\% $ for validation and testing over all the datasets.

\subsubsection{Baselines}
To evaluate the performance of \textsc{Self}$\mathcal{R}$\textsc{GNN}, we choose several state-of-the-arts baselines.
We only consider the models for temporal graphs as we are interested in the representation learning on temporal graphs in this study.

\noindent \textbf{Euclidean Model}:
\emph{For the supervised models}, we compare with the strong baselines including \textbf{VGRNN} \cite{2019Variational}, \textbf{EvolveGCN} \cite{pareja2019evolvegcn}, \textbf{DyRep} \cite{DyRep}, \textbf{TGAT} \cite{Xu2020inductive}  and the recent \textbf{CAWNet} \cite{CAWNet}. 
\emph{For the self-supervised model}, we include \textbf{DDGCL} \cite{DDGCL}, a recent contrastive learning method for temporal graphs.

\noindent \textbf{Riemannian Model}: \emph{For the supervised models}, we compare with the recent \textbf{HTGN} \cite{HTGN} and \textbf{HVGNN} \cite{HVGNN}, and both of them are in the Riemannian space of negative curvature (hyperbolic space). The proposed \textsc{Self}$\mathcal{R}$\textsc{GNN} is the first Riemannian model with time-varying curvature, to the best of our knowledge. \emph{For the self-supervised model}, there is few work in the literature, and we also fill this gap in \textsc{Self}$\mathcal{R}$\textsc{GNN}.

    \vspace{-0.05in}
\subsubsection{Evaluation Tasks}
Both node classification and link prediction are utilized as evaluation tasks.

\noindent \textbf{Node Classification}:
We evaluate the performance on Wikipedia, MOOC, and Cora. The node label of Wikipedia and MOOC is given following the study \cite{DDGCL}, and the nodes of Cora is given in the original citation network.
The  labels are utilized by the supervised models in both training and testing.
In contrast, similar to \citet{VelickovicFHLBH19}, self-supervised models first learn representations without labels, and then were evaluated by specific learning task, which is performed by directly using these representations to train and test.

\noindent \textbf{Link Prediction}:
We evaluate the performance on all the datasets.
In this work, we not only care about the link prediction between the trained nodes, but also expect the models to predict links between the new nodes. 
Hence, we introduce two types of link prediction tasks: 
i) \emph{Transductive link prediction} task allows temporal links between all nodes to be observed up to a time point during the training phase, and uses all the remaining links after that time point for testing. 
ii) \emph{Inductive link prediction} task predicts links associated with at least one node not observed in the training set. We first conduct the chronological data split as the transductive setting and then randomly select $10\%$ nodes to determine the edges to remove following the study \cite{CAWNet}. 

    \vspace{-0.05in}
\subsubsection{Implementation Details}  
To enhance the reproducibility, we provide the implementation details in the subsection.

\noindent \textbf{Euclidean Input}:
The input feature is Euclidean by default. In this case, we map input features to the Riemannian space before feeding into the non-Euclidean models. Specifically, we utilize the exponential map $exp _{\mathcal O}^{\kappa}(\cdot)$ to perform the mapping from $\mathbb R^d $ to $ \mathcal M^d_\kappa$.
% and thus we have $\boldsymbol{x}^\kappa$ as follows:
% \begin{equation}
%  \boldsymbol{x}^\kappa = \rho(\boldsymbol x)=\left[\frac{ \cos_\kappa \left(\sqrt{|\kappa|}\|  \boldsymbol x  \|_{\kappa}\right) }{\sqrt{|\kappa|}},
% \frac{ \sin_\kappa \left(\sqrt{|\kappa|}\|  \boldsymbol x  \|_{\kappa}\right) }{ \sqrt{|\kappa|}\|  \boldsymbol x  \|_{\kappa} } \boldsymbol x \right]
% \end{equation}
The curvature $\kappa$ can be either set as the parameter of a negative constant for the hyperbolic model, HVGNN and HGTN, or adopted the parametric formulation for the proposed \textsc{Self}$\mathcal{R}$\textsc{GNN}.

\noindent \textbf{\textsc{Self}$\mathcal{R}$\textsc{GNN}}: We stack the proposed temporal aggregation layers twice in \textsc{Self}$\mathcal{R}$\textsc{GNN}, and we utilize a two-layer MLP to build CurNN for curvature learning. The dimensionality of our temporal representations is set to $32$, while the hyperparameters of the baselines are set for the best performance according to the original papers.

 \begin{table}
    \centering
          \caption{Ablation study with node classification in terms of AUC (\%)  on Wiki, MOOC and Cora datasets.}
          \vspace{-0.1in}
    \begin{tabular}{c l | c c c}
      \toprule
       \multicolumn{2}{c|}{\textbf{Variant}}  &  \textbf{Wiki} & \textbf{MOOC} & \textbf{Cora}\\
     \toprule
     \multirow{2}{*}{ $\mathbb S$} 
       &  \textsc{Self}$\mathcal{R}$\textsc{GNN}$^+$w/oRR &  $ 85.18(0.3)$   &  $ 69.24(0.2)$   &  $ 83.60(0.2)$  \\
       &  \textsc{Self}$\mathcal{R}$\textsc{GNN}$^+$            &   $ 88.46(0.1)$   &  $ 73.78(0.6)$   &  $ 87.45(1.0)$  \\
       \midrule
      \multirow{2}{*}{ $\mathbb E$} 
      &  \textsc{Self}$\mathcal{R}$\textsc{GNN}$^0$w/oRR &    $ 83.95(0.1)$   &  $ 70.33(0.2)$   &  $ 84.17(0.5)$  \\
       &  \textsc{Self}$\mathcal{R}$\textsc{GNN}$^0$            &   $ 88.02(0.5)$   &  $ 75.82(0.2)$   &  $ 87.05(0.3)$  \\
       \midrule
        \multirow{2}{*}{ $\mathbb H$} 
      &  \textsc{Self}$\mathcal{R}$\textsc{GNN}$^-$w/oRR &    $ 85.60(0.6)$   &  $ 72.50(0.1)$   &  $ 88.33(0.1)$  \\
      &   \textsc{Self}$\mathcal{R}$\textsc{GNN}$^-$            &    $ 90.82(1.2)$   &  $ 78.16(0.5)$   &  $ \emph{\underline{93.89}}\ (0.3)$  \\
      \midrule 
      \multirow{3}{*}{ \rotatebox{90}{Varying }} 
      &   \textsc{Self}$\mathcal{R}$\textsc{GNN}w/oRR       &  $ 89.27(0.3)$      &  $ 77.08(0.3)$   &  $ 91.48(0.1)$  \\
      &   \textbf{\textsc{Self}$\mathcal{R}$\textsc{GNN}} &  $ \mathbf{93.64}(0.\  )$   &  $  \mathbf{81.28}(0.1)$   &  $ \mathbf{94.06}(0.2)$   \\
       \cline{2-5}
     &   $\mathcal{R}$\textsc{GNN}(Supervised)                  &   $ \emph{\underline{92.03}}\ (0.1)$   &  $ \emph{\underline{80.89}}\ (0.2)$   &  $ 92.50(1.1)$  \\
      \bottomrule
    \end{tabular} 
    \vspace{-0.05in}
        \label{results3}
  \end{table}

    \vspace{-0.05in}
\subsection{Overall Performance (\emph{RQ1})}

We utilize AUC, area under RoC curve, as the evaluation metric for the tasks of node classification and link prediction, and report its mean value with $95\%$ confidence interval of $10$ independent runs for each model to achieve fair comparisons.
The confidence interval are given in the brackets in the tables.

\vspace{-0.05in}
\subsubsection{Node Classification}
Traditional classifiers work with the Euclidean space, and cannot be directly applied to the Riemannian space due to the essential distinction in geometry. Thus, we first discuss the node classification in the Riemannian space. 
In this work, following \citet{HGNN}, we introduce an output transformation which transforms output representations to Euclidean encodings.
%we utilize the Euclidean encodings which summarize the structure of node representations for classification, similar to \cite{HGNN}.
Specifically, given an output $\boldsymbol h_i(t)$, we first introduce a set of centroids $\{\boldsymbol \mu_1,  \cdots, \boldsymbol \mu_C\}_{(t)}$,
where $\boldsymbol \mu_c(t)$ is the centroid in Riemannian space learned jointly with the learning model. %using backpropagation.
% Then, we transform the output representation $\boldsymbol z_j \in \mathcal M$ into an Euclidean encoding $\boldsymbol \xi\in \mathbb R^C$, 
% which summarizes the position of $\boldsymbol z_i $ relative to the centroids, i.e., $\boldsymbol \xi=\left(\xi_{1j}, \ldots, \xi_{Cj}\right) $ and $\xi_{ij}=d_{\mathcal M}(\boldsymbol \mu_i, \boldsymbol z_j)$.
Then,  encoding of $\boldsymbol h_i(t)$ is defined as $\boldsymbol \xi_i(t)=\left(\xi_{i1}, \ldots, \xi_{iC}\right)^\top $, where $\xi_{ij}=d_{\mathcal M}(\boldsymbol h_i(t), \boldsymbol \mu_j(t))$, summarizing the position of temporal representations relative to the centroids.
Now, we are ready to use logistic regression for node classification, and the likelihood is given as
\begin{equation}
p( y | \boldsymbol h(t) )=Sigmoid(\boldsymbol w^\top \boldsymbol h(t)),
\end{equation}
where $\boldsymbol w \in \mathbb R^{|C|}$ is the parameter, and $y$ is the label.
Let $\boldsymbol h(t)=\boldsymbol \xi(t)$ for non-Euclidean models (i.e., the hyperbolic HVGNN and HGTN, the proposed \textsc{Self}$\mathcal{R}$\textsc{GNN}) and $\boldsymbol h(t)$ is the output of Euclidean ones.
Note that, the hyperbolic logistic regression proposed in the study \cite{HNN} cannot be generalized to the Riemannian space of arbitrary curvature. 
We summarize the experimental results in Table \ref{results1}, and it is obvious that our \textsc{Self}$\mathcal{R}$\textsc{GNN} consistently outperforms the state-of-the-arts supervised methods.
The superiority  lies in that \textsc{Self}$\mathcal{R}$\textsc{GNN} model the temporal evolvement among hyperspherical, Euclidean, and hyperbolic spaces in reality, and the proposed self-supervised  approach learns temporal representations effectively. 

\begin{figure*} 
\centering 
\subfigure[The embedding space in May, 1996]{
\includegraphics[width=0.248\linewidth]{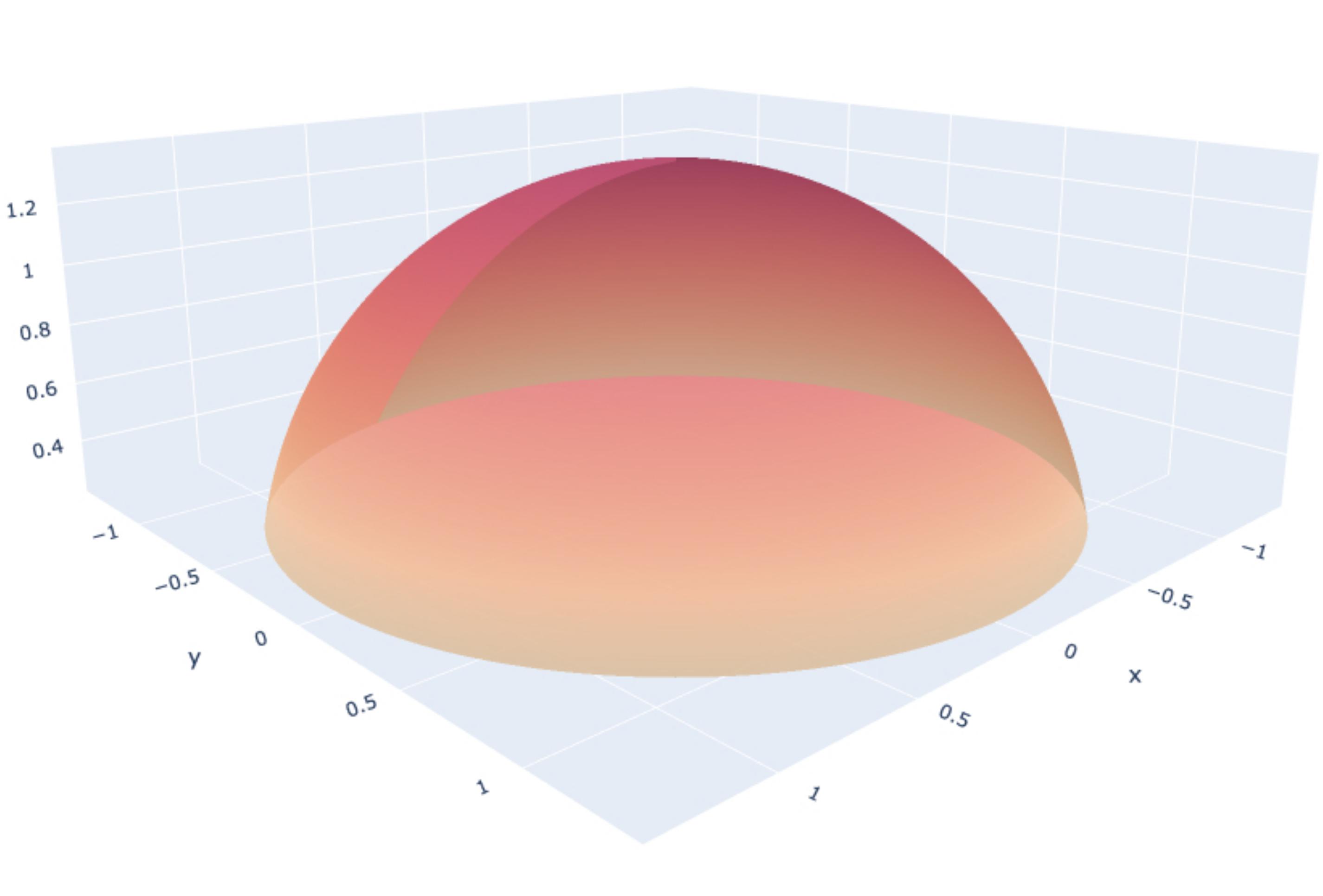}}
\hspace{-0.02\linewidth}
\subfigure[The embedding space in May, 1998]{
\includegraphics[width=0.248\linewidth]{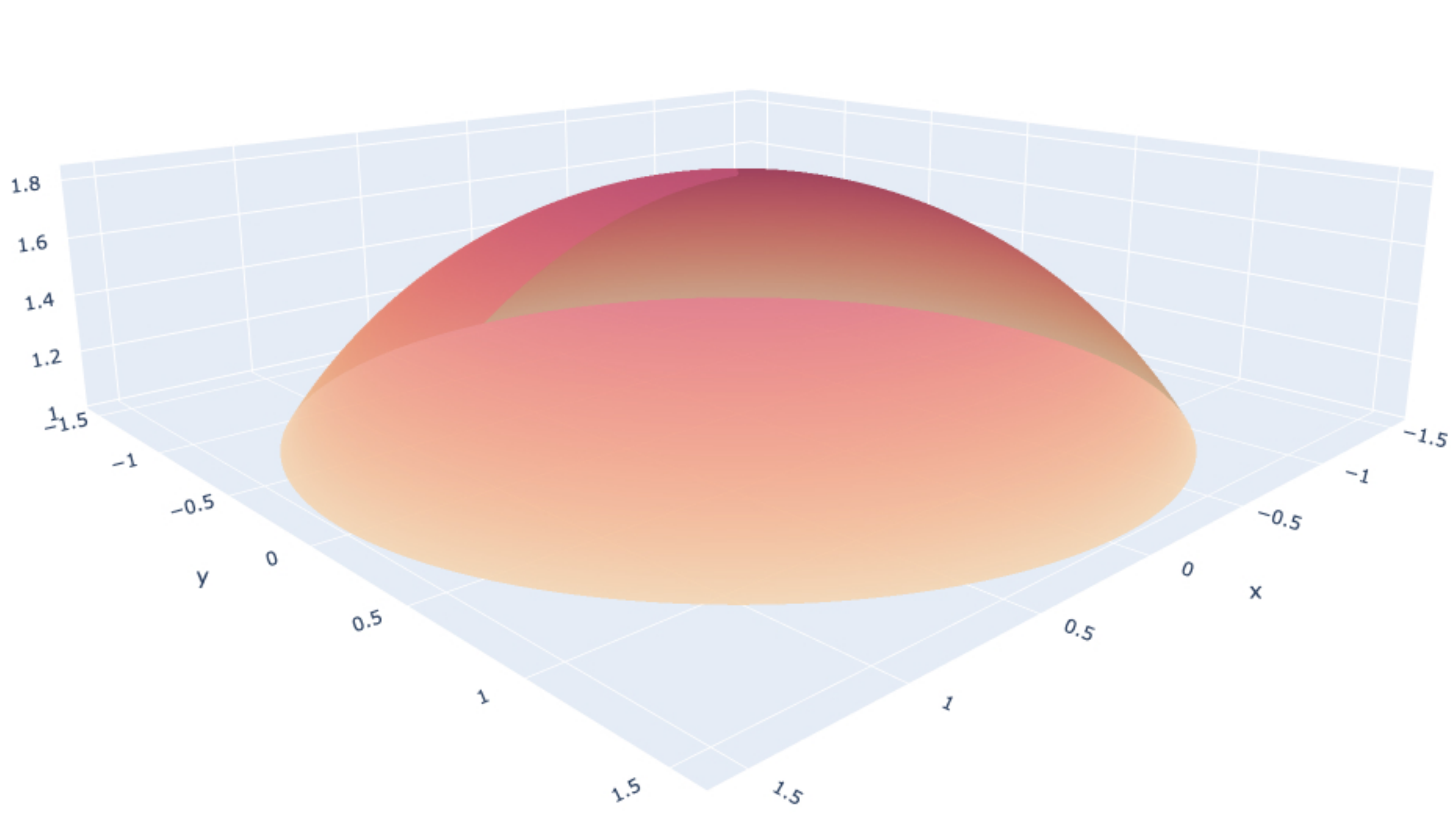}}
\hspace{-0.02\linewidth}
\subfigure[The embedding space in May, 2000]{
\includegraphics[width=0.248\linewidth]{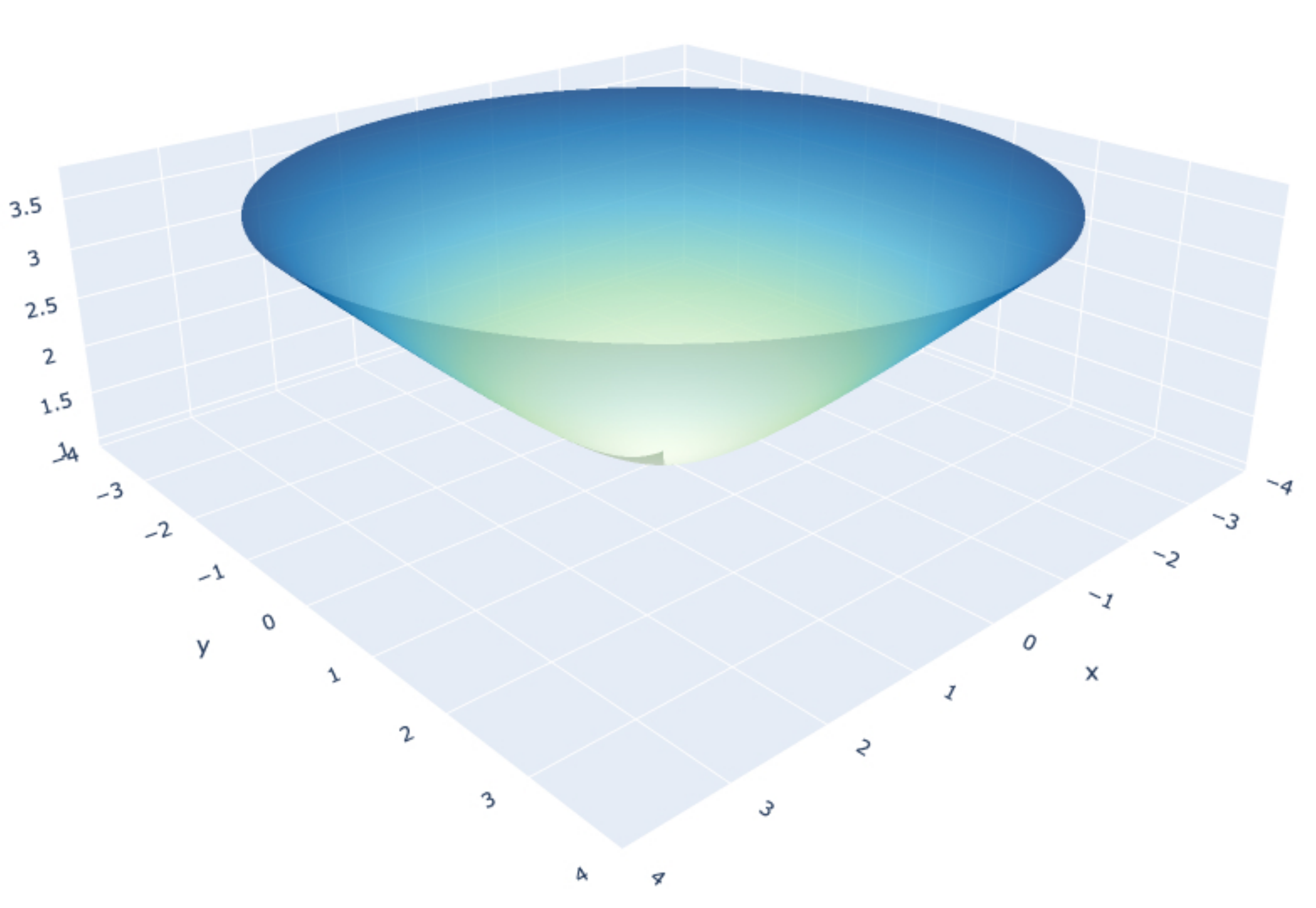}}
\hspace{-0.02\linewidth}
\subfigure[The embedding space in May, 2002]{
\includegraphics[width=0.248\linewidth]{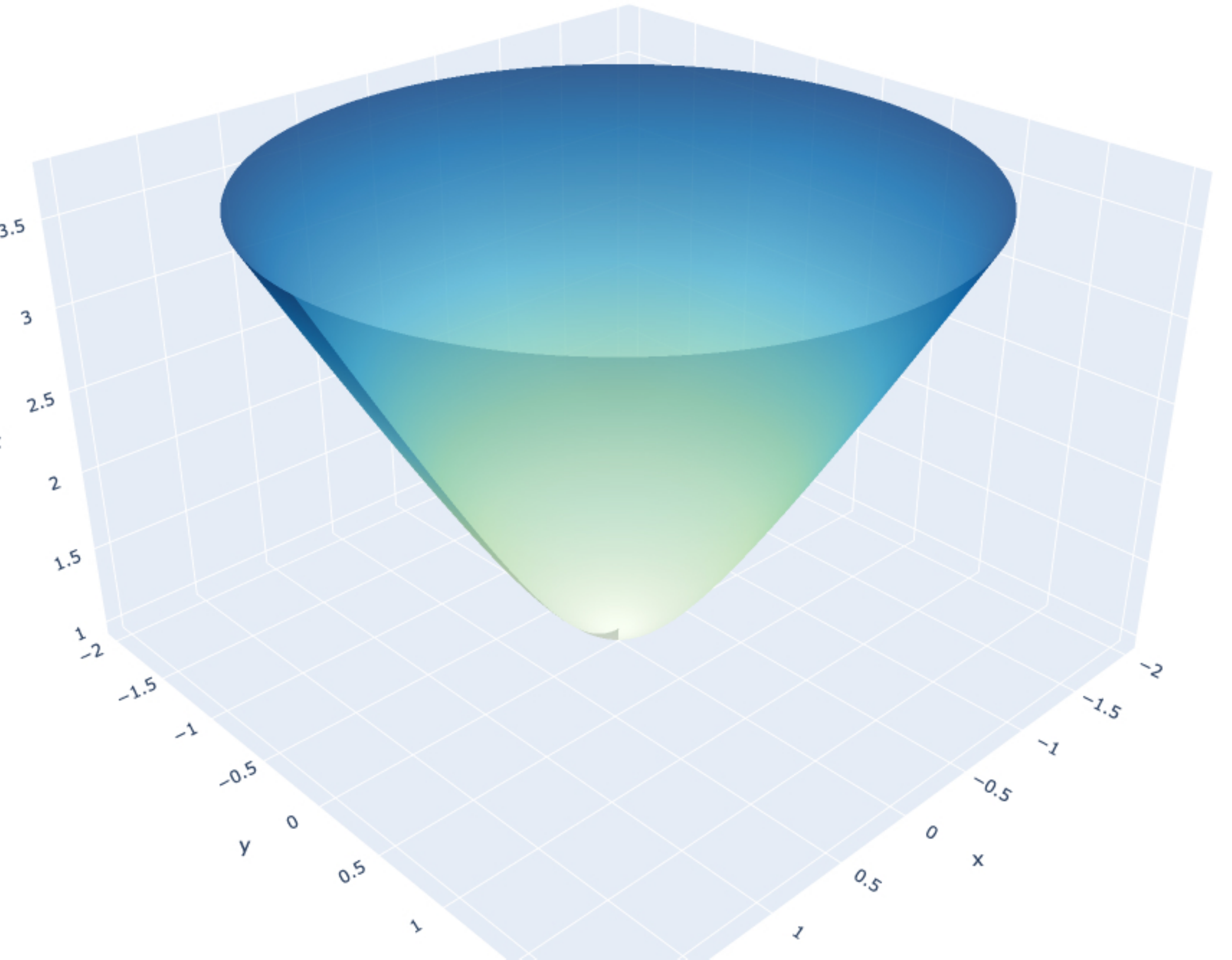}}
 \vspace{-0.18in}
\caption{Visualization of the embedding spaces over time on Physics datasets. }
 \vspace{-0.1in}
\label{curvature}
\end{figure*}

\subsubsection{Link Prediction}
For link perdition tasks, we utilize the Fermi-Dirac decoder, a generalization of sigmoid, to compute probability scores for edges.
Formally, given output representations $\boldsymbol h(t)$, we have the probability formulated as follows: % representations. 
\begin{equation}
\begin{aligned}
& p((v_i, v_j) \in \mathcal E|\  \boldsymbol h_i(t), \boldsymbol h_j(t)) = \left(\exp\left(\frac{d_\mathcal M(\boldsymbol h_i(t), \boldsymbol h_j(t))^2-r}{t}\right)+1\right)^{-1},
\end{aligned}
\end{equation}
where $r$ and $t$ are hyperparameters. 
For each method, $d_\mathcal M$ is the distance metric of corresponding representation space, e.g., $||\boldsymbol h_i(t) - \boldsymbol h_j(t)||_2$ for Euclidean models, and we utilize $d_\mathcal M$ in Table \ref{tab:ops} for the Riemannian models.
We report the experimental results of both inductive link prediction and transductive link prediction in Table \ref{results2}.
The proposed \textsc{Self}$\mathcal{R}$\textsc{GNN} achieves the best performance on all the datasets except for one case, the transductive setting on Wiki against CAWNet, which is a supervised method with a calibrated inductive bias for interaction prediction.
 % The inferiority to CAW-N is reasonable since the CAW model entails a carefully calibrated inductive bias for interaction prediction [43] and was previously shown to be strictly more expressive than the TGAT architecture on link prediction. However, the inductive bias brought by CAW may not be preferable in node classification settings, as illustrated in table 1.

 \vspace{-0.09in}
\subsection{Ablation Study (\emph{RQ2})}
We conduct ablation study to show how each proposed component contributes to the success of  \textsc{Self}$\mathcal{R}$\textsc{GNN}.
To this end, we design two types of variants as follows:
\begin{itemize}
\item The first type of variants is to verify the effectiveness of \emph{the time-varying curvature}. We use the superscript to distinguish the shape of Riemannian space, e.g., \textsc{Self}$\mathcal{R}$\textsc{GNN}$^-$ of the negative sign denotes the corresponding model work with the negative curvature hyperbolic space.

\item The second type of variants is to verify the effectiveness of the \emph{\underline{R}iemannian \underline{R}eweighed contrastive loss}. We use the suffix of \textbf{w/oRR} to denotes the corresponding model trained with the original InfoNCE loss.
\end{itemize}

\noindent Additionally, we train the curvature-varying $\mathcal{R}$\textsc{GNN} with labels, referred to as $\mathcal{R}$\textsc{GNN}(Supervised), to show the effectiveness of the proposed self-supervised approach. 
That is, we have eight variants in total, and note that \textsc{Self}$\mathcal{R}$\textsc{GNN}$^0$ means that the proposed \textsc{Self}$\mathcal{R}$\textsc{GNN} degenerates into a special case of Euclidean space. 
We utilize node classification as the evaluation task for the ablation study, and the performance of variants are summarized in Table \ref{results3}.
We find that: 
i) The proposed \textsc{Self}$\mathcal{R}$\textsc{GNN} with time-varying curvature outperforms its variants with a certain curvature, especially for the zero curvature (Euclidean space). We will further discuss it in our case study.
ii) The performance of the proposed RR loss beats that of the original InfoNCE loss, and \textsc{Self}$\mathcal{R}$\textsc{GNN} even obtain better results than the supervised $\mathcal{R}$\textsc{GNN}, showing the effectiveness of Riemannian reweighted contrastive loss.

\subsection{Case Study and Discussion (\emph{RQ3})}
In the case study, we show the curvature of the underlying geometry evolves over time on Physics  \cite{2019Variational}, a citation network related to high energy physics from Jan, 1993 to April, 2003.
We embed the network at different time points in the $2$-dimensional Riemannian spaces, and learn the corresponding curvatures from the data, illustrated in Fig. 3, where the curvature is reported in Table \ref{results4}.
As shown in our case study, \emph{rather than remained in a certain curvature,  the underlying space  evolves from positive curvature hyperspherical to negative curvature hyperbolic space}.
At the beginning, a number of paper join in the network forming triangles or other cyclical structures.
%, which presents a positive curvature of hyperspherical spaces.
As the time progresses, high-impact papers of high citations acquire a better visibility to receive more citations, thus making the underlying geometry evolve to be hyperbolic.
%than the low-impact ones.
%high-impact works trend to receive attentions more easily than the low-impact ones, and have more citations.
Similar phenomenon is also observed in the study \cite{ravasz2003hierarchical}.
% Correspondingly, the graph presents hierarchical structure, and 
That is, the embedding space shifts among the Riemannian space of various curvatures (positive, zero and negative) in the graph evolvement over time, 
explaining the inferior of methods with certain curvature and the superior of the proposed \textsc{Self}$\mathcal{R}$\textsc{GNN} with time-varying curvatures.

  \begin{table}
    \centering
          \caption{Curvature evolvement on Physics datasets.}
          \vspace{-0.1in}
        \begin{tabular}{p{1.5cm}<{\centering} |  p{1.3cm}<{\centering} p{1.3cm}<{\centering}  p{1.3cm}<{\centering} p{1.3cm}<{\centering}}
     \toprule
    % Time point $(t/T)$ & $20\%$ & $40\%$ & $60\%$ & $80\%$ \\
     Time point  & May, 1996 &May, 1998 & May, 2000& May, 2002 \\
      \midrule
      Curvature  & $\emph{+0.552}$ & $\emph{+0.303}$  & $\emph{-0.259}$ & $\emph{-1.022}$ \\
      \bottomrule
    \end{tabular} 
        \label{results4}
                  \vspace{-0.15in}
  \end{table}

\vspace{-0.07in}
\section{Related Work}

Our work mainly relates to temporal graph learning and Riemannian representation learning.
\vspace{-0.07in}
\subsection{Temporal Graphs Learning}
Representation learning on temporal graphs \cite{AggarwalS14,KazemiGJKSFP20} consider the representations to be time-dependent as the graph evolves over time.
% , which can be roughly divided into two main categories: discrete-time methods and continuous-time methods. 
% The discrete-time methods operate on a sequence of snapshots, while the continuous-time methods directly model the temporal interactions.  
For the discrete-time methods, recurrent architectures are frequently employed capture the time-dependence over snapshots, e.g.,
VGRNN \cite{2019Variational} and EvolveGCN \cite{pareja2019evolvegcn}.
For the continuous-time methods, temporal random walks \cite{CTDNE,liu2020fine} have shown to be effective, and the recent CAWNet \cite{CAWNet} is based on the causual anonymous walks.
%to learn the temporal representations inductively
Temporal point process \cite{TREND,zuo2018embedding,HeteroHawkes} is another important tool, e.g., DyRep \cite{DyRep} considers an additional hop of interactions for further expressiveness. 
Recently, GNN-based models have also emerged to deal with continuous time, e.g., TGAT \cite{Xu2020inductive} extends GAT \cite{velickovic2018graph} to the temporal graphs. 
%gives an elegant way 
JODIE \cite{JOIDE} models the message exchanges with the mutual RNNs for bipartite graphs specifically.
These studies usually rely on the labels to learn the representations.
% In fact, the self-supervised methods are more preferable for the unlabeled graphs of real applications, but unfortunately, the effort for the self-supervised learning on temporal graphs is still limited.
Recently, DDGCL \cite{DDGCL} enables the self-supervised learning in the traditional Euclidean space as the prior works do.
To the best of our knowledge, none of the existing studies consider \emph{the self-supervised learning on temporal graphs  in the general Riemannian space}. 

\subsection{Riemannian Representation Learning}

Recently, it emerges as an exciting alternative to the traditional Euclidean representation learning \cite{nickel2017poincare,mathieu2019continuous,HNN,DBLP:conf/icdm/0008Z0WDSY20}.
In this subsection, we mainly focus on the Riemannian representation learning on \emph{graphs}.
% Recent advances of network science \cite{krioukov2010hyperbolic} show that  hyperbolic space, the Riemannian space of negative-curvature, is well-suited to model the graphs with latent hierarchical or tree-like structures, and thus motivates the graph representation learning in hyperbolic space.
Most of the existing studies in the literature investigate on the static graphs.
A number of hyperbolic GNNs are proposed, e.g., HAN \cite{HAN} generalize the attention mechanism. HGCN \cite{HGCN}, HGNN \cite{HGNN} and LGNN \cite{ZhangWSLS21} design the hyperbolic graph convolution with different formulations.
\citet{DBLP:conf/icdm/Fu0WSJWTPY21} studies how to select the optimal curvature for hyperbolic GNN in a joint optimization objective with a reinforcement learning method.

Beyond the hyperbolic space, $\kappa$-GCN  \cite{BachmannBG20} generalizes the GCN to arbitrary constant-curvature spaces with the $\kappa$-sterographical model. 
\citet{DBLP:conf/www/YangCPLYX22} propose to represent graphs in the dual space of Euclidean and hyperbolic ones.
Recently, \citet{SelfMGNN} propose to study graph learning in the mixed-curvature space, and enable the self-supervised learning with a novel Riemannian contrastive learning. 
A concurrent work introduces a mixed-curvature model for knowledge graphs specifically with the supervised fashion \cite{DBLP:conf/www/WangWSWNAXYC21}.
It is not until very recently the Riemannian representation learning for temporal graphs are explored.
Concretely, HVGNN \cite{HVGNN} directly models the temporal interaction underpinned by a novel invariant encoding with an attentional architecture in the Lorentz model.
In the meanwhile, HTGN \cite{HTGN} designs a novel recurrent architecture on the sequence of snapshots to incorporate the temporal information on the graphs.
Both of them restrict themselves in the negative curvature hyperbolic space. 
%and require external guidance to learn the representation.
\emph{Distinguishing with the prior works, we propose the first time-varying curvature model shifting among hyperspherical, Euclidean, and hyperbolic spaces in the evolvement}.

%generalize the negative-curvature Riemannian space to a Riemannian space of arbitrary curvature, and propose \emph{the first self-supervised temporal graph neural network in Riemannian space}.

\vspace{-0.15in}
\section{Conclusion}

In this paper, we for the first time study the representation learning problem on temporal graph in the general Riemannian space, shifting among hyperspherical, Euclidean, and hyperbolic spaces in the evolvement.
To this end, we present the novel \textsc{Self}$\mathcal{R}$\textsc{GNN}.
Specifically, we propose the  curvature-varying Riemannian GNN with the theoretically grounded time encoding, so that we can design the functional curvature to model the evolvement over time. 
To explore the rich information in the Riemannian space itself, we propose the reweighting self-contrastive approach for representation learning and the edge-based self-supervised curvature learning with  Ricci curvature.
Extensive experiments on real-world temporal graphs show the superiority of \textsc{Self}$\mathcal{R}$\textsc{GNN}, and in addition, we study the curvature evolvement over time on temporal graphs.
%against the state-of-the-arts methods.

%explore the rich information of the temporal graphs in the Riemannian space themselves without the effort for graph augmentation, 
% \vspace{-0.1in}
% \begin{acks}
% % To Robert, for the bagels and explaining CMYK and color spaces.
% This paper was supported in part by 
% National Key R\&D Program of China through grant 2021YFB1714800,
% the Fundamental Research Funds for the Central Universities 2022MS018 and
%  S\&T Program of Hebei through grant 21340301D.
% Philip S. Yu is supported in part by NSF under grants III-1763325, III-1909323,  III-2106758, and SaTC-1930941. 
% For any correspondence, please refer to Li Sun and Hao Peng.
% \end{acks}

\bibliographystyle{ACM-Reference-Format}
\bibliography{cikm22}

% \appendix









\end{document}